\documentclass[sn-mathphys-ay]{sn-jnl}

\usepackage{graphicx}%
\usepackage{multirow}%
\usepackage{amsmath,amssymb,amsfonts}%
\usepackage{amsthm}%
\usepackage{mathrsfs}%
\usepackage[title]{appendix}%
\usepackage{xcolor}%
\usepackage{textcomp}%
\usepackage{manyfoot}%
\usepackage{booktabs}%
\usepackage{algorithm}%
\usepackage{algorithmicx}%
\usepackage{algpseudocode}%
\usepackage{listings}%
\usepackage{array}
\usepackage[caption=false,font=footnotesize]{subfig}
\usepackage{hyperref}

\begin{document}

\title[Article Title]{Learning to Learn for Few-shot Continual Active Learning}


\author*[1,2]{\fnm{Stella} \sur{Ho}}\email{stella.ho@unimelb.edu.au}

\author*[1]{\fnm{Ming} \sur{Liu}}\email{m.liu@deakin.edu.au}

\author[1]{\fnm{Shang} \sur{Gao}}\email{shang.gao@deakin.edu.au}

\author[3]{\fnm{Longxiang} \sur{Gao}}\email{gaolx@sdas.org}

\affil[1]{\orgdiv{School of Information Technology}, \orgname{Deakin University}, \orgaddress{\city{Burwood}, \postcode{3125}, \state{Victoria}, \country{Australia}}}

\affil[2]{\orgdiv{Department of Biomedical Engineering}, \orgname{The University of Melbourne}, \orgaddress{\city{Parkville}, \postcode{2503}, \state{Victoria}, \country{Australia}}}

\affil[3]{\orgdiv{Computer Science Center}, \orgname{Qilu University of Technology}, \orgaddress{\city{Jinan}, \postcode{250353}, \state{Shandong}, \country{China}}}


\abstract{Continual learning strives to ensure \textit{stability} in solving previously seen tasks while demonstrating \textit{plasticity} in a novel domain. Recent advances in continual learning are mostly confined to a supervised learning setting, especially in NLP domain. In this work, we consider a few-shot continual active learning setting where labeled data are inadequate, and unlabeled data are abundant but with a limited annotation budget. We exploit meta-learning and propose a method, called \textit{Meta-Continual Active Learning}. This method sequentially queries the most informative examples from a pool of unlabeled data for annotation to enhance task-specific performance and tackle continual learning problems through meta-objective. Specifically, we employ meta-learning and experience replay to address inter-task confusion and catastrophic forgetting. We further incorporate textual augmentations to avoid memory over-fitting caused by experience replay and sample queries, thereby ensuring generalization. We conduct extensive experiments on benchmark text classification datasets from diverse domains to validate the feasibility and effectiveness of meta-continual active learning. We also analyze the impact of different active learning strategies on various meta continual learning models. The experimental results demonstrate that introducing randomness into sample selection is the best default strategy for maintaining generalization in meta-continual learning framework.}




\maketitle

\section{Introduction}\label{sec1}

Continual Learning (CL) aims to address the \textit{stability-plasticity} dilemma in the context of sequential learning. \textit{Stability} refers to the ability to alleviate the decline in model performance on previously learned tasks, in other words, preventing catastrophic forgetting \citep{McCloskey1989CatastrophicII}. Notably, inter-task confusion \citep{DBLP:conf/aaai/HuangC0CW23} is one of the major reasons that causes catastrophic forgetting. It is a phenomenon where the model confuses classes from different tasks. \textit{Plasticity} refers to the ability to adapt to a new task. However, the majority of existing CL methods presume the availability of labeled data in abundance, deeming it adequate for learning every task. Their performance heavily relies on the large quantity and high quality of labeled data, which is impractical. 

In real-world scenarios, labeled data are scarce while unlabeled data are abundant. The cost of annotation, particularly in the field of Natural Language Processing (NLP), tends to be prohibitively high. To narrow this gap, we address the problems in a setting that more closely aligns with real-world scenarios, namely few-shot \textit{Continual Active Learning} (CAL) \citep{DBLP:conf/nips/AyubF22}. In this setting, only a small subset of labeled data is provided for each task with a limited annotation budget. In such a way, model should sequentially select the most worthwhile examples from a pool of unlabeled data and requests labels to enhance performance and simultaneously solve continual learning problems. Introducing active learning into the scheme can be challenging as active learning techniques are typically designed to query from a static data distribution. They may not be able to dynamically capture most relevant samples for queries to prevent catastrophic forgetting.

Replay-based methods have been shown to be particularly effective for NLP tasks \citep{wang-etal-2022-learning-robust}. These methods often retain a certain amount of past samples to prevent catastrophic forgetting.  Consequently, they are prone to memory over-fitting, especially when the availability of labeled examples is limited. However, active learning can further escalate the problem by biasing the sample selection towards the memory set. Hence, integrating active learning techniques into CL model can be quite challenging.

Given the success of meta-learning on solving low plasticity, certain studies \citep{DBLP:conf/iclr/RiemerCALRTT19,10.5555/3495724.3496696} extend meta-learning to CL setting. In this work, we exploit the advantages of meta-learning and uses Model-Agnostic Meta-Learning (MAML) \citep{DBLP:conf/icml/FinnAL17} framework to solve few-shot CAL problems, called \textit{Meta-Continual Active Learning}. By applying active learning for task-specific tuning and casting meta-objective as experience replay, the learning objective is deliberately formulated to learn an optimal or a suboptimal initial model state that can rapidly adapt to a balanced subset of all encountered task. Thereby, this method enables fast adaptation while preventing catastrophic forgetting in few-shot learning. In addition, we apply consistency regularization via textual augmentations to address memory overfitting problems that inherent in replay-based method and exacerbated by active learning acquisition.

We conduct extensive experiments on benchmark datasets from \citep{DBLP:conf/nips/ZhangZL15}, popularized by \citep{DBLP:conf/nips/dAutumeRKY19} in lifelong language learning. This collection of datasets includes five text classification datasets from four diverse domains. We demonstrate the effectiveness of the proposed framework in a 5-shot CAL setup. This paper also examines how various active learning approaches impact the performance of meta continual learning models.

The main contributions of this paper are fourfold:

\begin{itemize}    
    \item  Leveraging the strengths of meta-learning, we introduce an optimization-based method, namely \textit{Meta-Continual Active Learning} (Meta-CAL). This method reformulates the meta-objective such that it learns an optimal or a suboptimal initial model state that can effectively adapt to all seen tasks. Thereby, it provides a solution to inter-task confusion and catastrophic forgetting even with limited availability of labeled samples. 
    
    \item We integrate active learning into the proposed framework to enhance task-specific tuning. This allows the model to dynamically and selectively query the most informative samples from a pool of unlabeled data, thereby improving performance in a resource-constrained scenario.

    \item To address inevitable memory overfitting problems caused by experience replay and active learning, we apply consistency regularization to meta examples via data augmentations. This further ensures intra- and inter-task generalization.
    
    \item In the experiments, the proposed method achieves an accuracy of more than 62\% while utilizing only 1.6\% of the past samples and maintaining annotation budgets as low as 500 samples for each task. The results demonstrate the feasibility and effectiveness of meta-continual active learning. Furthermore, we observe that random sampling facilitates generalization in meta continual learning, thereby addressing the stability-plasticity dilemma.

\end{itemize}

\section{Related Work}\label{sec:related_work}
\subsection{Continual Learning}
Existing approaches can be categorised into three main mainstreams, i.e., regularization-based methods \citep{Kirkpatrick2017OvercomingCF,Li2018LearningWF}, replay-based methods \citep{DBLP:conf/nips/dAutumeRKY19,10058177} and architecture-based methods \citep{DBLP:conf/iclr/Adel0T20, DBLP:conf/iclr/YoonYLH18}. 

Regularization-based methods often impose a penalty or regularization term to the loss function. These methods often penalise changes in non-trivial parameters or constrain the variations in gradients learned from previous tasks. However, most methods for NLP show a preference for replay-based approaches \citep{wang-etal-2022-learning-robust} to avoid unexpected output from tuning the parameters of deep neural networks \citep{DBLP:conf/naacl/WangXYGCW19}. 

Replay-based methods, also known as rehearsal-based methods or memory-based methods, involves revisiting a small amount of past samples (i.e., experience replay) or generating pseudo past samples while adapting to new domain. The popular retrieval schemes for experience replay are random \citep{DBLP:conf/iclr/ChaudhryRRE19,DBLP:conf/iclr/RiemerCALRTT19,DBLP:conf/nips/dAutumeRKY19,Holla2020MetaLearningWS}, K-Means \citep{DBLP:conf/naacl/WangXYGCW19,DBLP:conf/acl/HanDGLLLSZ20} and Mean-of-Feature \citep{DBLP:conf/acl/QinJ22,DBLP:conf/acl/ChenWS23}.

Architecture-based methods dynamically change the model architecture to learn a new task. In general, these methods preserve or partially preserve past fine-tuned parameters and introduce task-specific parameters for the new domain. However, it is challenging to manage the scale of the model since the parameter size continuously accumulates with the increase in the number of seen tasks.

\subsection{Meta Continual Learning}
In recent years, meta-learning has emerged as an effective learning framework for CL. In particular, the bi-level optimization of meta-learning enables fast adaptation to training data while ensuring generalization across all observed samples. In meta-continuous learning, meta-learning is often combined with memory replay. Meta-MbPA \citep{DBLP:conf/emnlp/WangMPC20} uses MAML to augment episodic memory replay via local adaptation. OML-ER \citep{Holla2020MetaLearningWS} and ANML-ER \citep{Holla2020MetaLearningWS} utilise an online meta-learning model (OML) \citep{DBLP:conf/nips/JavedW19} and a neuromodulated meta-learning (ANML) \citep{DBLP:conf/ecai/BeaulieuFMLSCC20} respectively for effective knowledge transfer through fast adaptation and sparse experience replay. PMR \citep{10058177} also employs MAML for facilitating episodic memory replay but using a prototypical memory sample selection approach. MER \citep{DBLP:conf/iclr/RiemerCALRTT19} regularizes the objective of experience replay by graident alignment between old and new tasks via a modified Reptile \citep{DBLP:journals/corr/abs-1803-02999}. C-MAML\citep{10.5555/3495724.3496696} utilizes OML to regulate CL objectives and La-MAML\citep{10.5555/3495724.3496696} optimizes OML objective through the modulation of per-parameter learning rates. Meta-CL\citep{wu2024meta} further improves C-MAML and La-MAML by introducing penalty to restrain unnecessary model updates and preserve non-trivial weights for knowledge consolidation. To date, not many models use gradient alignment for continual learning in NLP. In addition, all of these models are employed in a supervised setting. In this work, we focus on aligning meta-learning with CL objectives to provide a solution in a realistic and resource-constrained scenario.

\subsection{Continual Active Learning} 
Continual active learning aims to sequentially labels informative data to maximise model performance and solve continual learning problems. It defines a continual learning problem where the available labeled data are insufficient and the annotation budgets are limited. CASA \citep{Perkonigg2021ContinualAL} detects new pseudo-domains and selected data from new pseudo-domains for annotation, while it revisits labeled samples to address catastrophic forgetting. \cite{DBLP:conf/nips/AyubF22} propose a method to address few-shot CAL (FoCAL). They use a Guaussian mixture model (GMM) for active learning and pseudo-rehearsal for CL, bypassing the need to store real past data. However, neither of these methods addresses continual active learning in NLP. CAL-SD \citep{das2023continual} tackles NLP tasks, and uses model distillation to augment memory replay with a diversity and uncertainty AL strategies. To date, continual active learning remains understudied, especially in NLP.

\section{Preliminaries} \label{sec:pre}
In this work, we focus on task-free class-incremental learning scenario, where training data stream passes only once without the presence of ``task boundaries'' \citep{DBLP:conf/cvpr/Ven0T21}. Based on the task-free setting, we formulate the problem of few-shot continual active learning as follows. Assume that training stream consists of $T$ tasks, $\{\mathcal{T}_1,\mathcal{T}_2,...,\mathcal{T}_t,...,\mathcal{T}_T\}$. Each task $\mathcal{T}_{t}$ contains $N_t$-way $K$-shot set $\mathcal{D}^{label}_t = \{(x_{i},y_{i})\}_{i=1}^{N_t \times K}$ and a pool of unlabeled data $\mathcal{D}^{pool}_t = \{(u_{i})\}_{i=1}^{|\mathcal{D}^{pool}_t|}$.

\subsection{Label Space}
Based on the label space $\mathcal{Y}$ of tasks, typical continual learning scenarios have \textit{domain-incremental learning}, where $\mathcal{Y}_t = \mathcal{Y}_t', \forall t \neq t'$ and \textit{class-incremental learning}, where $\mathcal{Y}_t \cap  \mathcal{Y}_t'= \emptyset, \forall t \neq t'$.  Due to the unforeseen nature of sequential learning, the label space $\mathcal{Y}_t$ of a task may or may not be disjoint from the other tasks. Hence, we allow $\mathcal{Y}_t \cap  \mathcal{Y}_t' \neq \emptyset$ or $\mathcal{Y}_t \cap  \mathcal{Y}_t'= \emptyset, \forall t \neq t'$ to occur.

\subsection{Annotation Constraint}
We consider each task $\mathcal{T}_{t}$ to have an equal annotation budget $B_{A}$. An acquisition function $a(\cdot)$ dynamically query informative data points for annotation until acquisition process reaches annotation budget. We denote the selected samples for annotation as $a(\mathcal{D}^{pool}_t, B_{A})$ and newly annotated sample set as $\mathcal{D}^{new}_t$. 

\subsection{Memory Constraint}
Replay-based model revisits a small subset of labeled data $\mathcal{D}_{1:1-t}$ to regularize model $f$ while learning $\mathcal{T}_t$. We limit the amount of past training data saved in memory buffer $\mathcal{M}$, which should not exceed memory budget threshold $B_{\mathcal{M}}$.

\subsection{Objectives} 
Assume we have a learner $f_{\theta}$ and current task $\mathcal{T}_t$, the learning objectives are: (a) perform adaptation on $\mathcal{D}_t = \mathcal{D}^{label}_t \bigcup \mathcal{D}^{new}_t$, i.e., \textit{plasticity}:
\begin{equation} \label{eqn:cl_step_1}
    \tilde{\theta}_{t} = \arg \min_{\theta_{t} \in \Theta} \mathbb{E}_{ (x,y) \sim \mathcal{D}_{t}} [\mathcal{L}\big(f_{\theta_t}(x),y \big)]
\end{equation} where $\mathcal{L}_{\mathcal{T}_t}$ is the task loss, $\theta_{t}$ is the initial state and $\theta_{t} = \tilde{\theta}_{t-1}$; (b) prevent inter-task confusion and catastrophic forgetting of prior tasks, i.e., \textit{stability}:
\begin{equation}\label{eqn:cl_obj}
    \min_{\tilde{\theta}_t} \frac{1}{|t-1|} \sum_{i=1}^{t-1} \mathbb{E}_{ (x,y) \sim \mathcal{D}_{i}} [\mathcal{L}\big(f_{\tilde{\theta}_t}(x),y \big)] 
\end{equation}

\subsection{Active Learning Strategies} 
In this work, we consider four popular active learning (AL) methods as follows.

\subsubsection{Uncertainty} 
This method samples data $\boldsymbol{x}$ with high uncertainty. The uncertainty is measured by model outputs $\hat{y}$. \textit{Least-confidence} method \citep{DBLP:conf/aaai/CulottaM05} evaluates uncertainty by the confidence in predication, where lowest posterior
probability indicates greater uncertainty $\alpha_{LC}(\boldsymbol{x}, n_a)= -Pr(\hat{y}|\boldsymbol{x})$. \textit{Margin-confidence} \citep{37648} method considers the confidence margin between two most likely predictions $(\hat{y}_1, \hat{y}_2)$, $\alpha_{Marg.}(\boldsymbol{x}, n_a)= -|Pr(\hat{y}_1|\boldsymbol{x})-Pr(\hat{y}_2|\boldsymbol{x})|$. Small difference indicates high uncertainty. \textit{Entropy-based} \citep{10.1145/584091.584093} method uses predictive entropy $H$ as the indicator, where higher entropy shows more uncertainty in the posterior probability $\alpha_{Entr.}(\boldsymbol{x}, n_a)= H(Pr(\hat{y}|\boldsymbol{x}))$.

\subsubsection{Representative} 
This method selects data $\boldsymbol{x}$ that are geometrically representative in vector space \citep{DBLP:conf/acl/SchroderNP22}. In this work, input data $\boldsymbol{x}$ that have the shortest euclidean distance to the centroid of a cluster are representative. \textit{KMeans} uses the centroid of each cluster, which applies unsupervised learning to partition the data into clusters. The number of clusters is the selection size $n_a$. In this work, we also introduce \textit{Mean vectors} method as a baseline for comparison. It averages representation vectors of each training batch as the centroid.

\subsubsection{Diversity} 
The method chooses data $\boldsymbol{x}$  that are geometrically seen as outliers in vector space \citep{mosqueira2022human}. In this work, input data $\boldsymbol{x}$ with the longest Euclidean distance from a centroid are considered diverse. The centroid selections align with those used in the Representative method. 

\subsubsection{Random} 
This method randomly samples data $\boldsymbol{x}$ from unlabeled pool $\mathcal{D}^{pool}$.

\section{Learning to Learn for CAL} \label{sec:method}
Model-Agnostic Meta-Learning (MAML) \citep{DBLP:conf/icml/FinnAL17} is an optimization-based meta-learning approach, often referred to as the ``learning-to-learn'' algorithm.  Learning to learn allows a model to adapt to different data distributions, which can be seen as a form of transfer learning that improves generalization \citep{DBLP:conf/nips/AndrychowiczDCH16}. Therefore, we exploit the MAML framework to facilitate knowledge transfer and generalization across tasks. Moreover, we harness its fast adaption ability to address the challenge of resource scarcity.

\subsection{Learning to Fast Adapt} 
We approximately align the meta-objective with the objectives shown in Eqn. \ref{eqn:cl_step_1} and Eqn. \ref{eqn:cl_obj} as follows,
\begin{align}
  \min_{\theta_{t}}  \quad & \frac{1}{|t|} \sum_{i=1}^{t} \mathbb{E}_{ (x,y) \sim \mathcal{D}_{i}} [\mathcal{L}\big(f_{\tilde{\theta}_t}(x),y \big)] \\
  \text{s.t.} \quad & \tilde{\theta}_t = U_{\mathcal{T}_t, \mathcal{D}_t}(\theta_{t}) \nonumber
\end{align} where $U_{\mathcal{T}_t, \mathcal{D}_t}(\theta_{t})$ is update operation on $\theta_{t}$ using training set $\mathcal{D}_t$. Specifically, $U_{\mathcal{T}_t, \mathcal{D}_t}(\theta_{t})$ describes optimization acting on $\theta_{t}$ as
\begin{equation}
     U_{\mathcal{T}_t, \mathcal{D}_t}(\theta_{t}) = \theta_t - \alpha \nabla_{\theta_t} \mathbb{E}_{(x,y) \sim \mathcal{D}_t}\mathcal{L} (f_{\theta_t}(x),y) 
\end{equation} Hereby, instead of finding tuned $\tilde{\theta}_t$, model $f$ learns an optimal initialization $\theta_{t}$ that can effectively adapt to $\mathcal{D}_{1:t}$ with few labeled examples.

\subsection{Learning to Continually Learn} 
In general, the full datasets $\mathcal{D}_{1:t-1}$ are not available while learning $\mathcal{D}_{t}$. Thus, we retrieve a small subset of past samples $\mathcal{M}$ for experience replay in meta-objective,
\begin{align} \label{eqn:cal_obj}
  \min_{\theta_{t}} \quad  &  \sum_{(x,y) \in \mathcal{M}} [\mathcal{L}\big(f_{\tilde{\theta}_t}(x),y \big)] \\
  \text{s.t.} \quad & \tilde{\theta}_t = U_{\mathcal{T}_t, \mathcal{D}_t}(\theta_{t}) \nonumber \\
     & = \theta_t - \alpha \nabla_{\theta_t} \mathbb{E}_{(x,y) \sim \mathcal{D}_t}\mathcal{L} (f_{\theta_t}(x),y) \nonumber
\end{align} In such a way, we leverage the selected examples from the past to constrain the learning behaviour of $f$. As a result, it effectively tackles the problems of  \textit{inter-task confusion} and \textit{catastrophic forgetting}.

\subsection{Learning to Generalize} 
We also exploit data augmentations to enhance model's ability to generalize. Especially, we apply consistency regularization \citep{DBLP:conf/nips/BachmanAP14}, which is built on the assumption that perturbations of the same input should not affect the output. Inspired by FixMatch \citep{DBLP:conf/nips/SohnBCZZRCKL20}, we employ two types of data augmentations in meta-training, strong and weak, denoted by $\mathcal{A}(\cdot)$ and $\alpha(\cdot)$, respectively. In contrast to FixMatch, we employ textual augmentations under full supervision to ensure generalization with limited data availability. Specifically, weak and/or strong augmentations are applied in the inner loop to enhance \textit{intra-task generalization}, while strong augmentations are used in the outer loop to improve both \textit{intra-} and \textit{inter-task generalization}. More details are shown in \S \ref{subsec:train}.

\section{Model} \label{sec:model}
\subsection{Model Architecture} 
Following Online-aware Meta-Learning (OML) \citep{DBLP:conf/nips/JavedW19}, the proposed model $f_{\theta}$ consists of a representation learning network $h_{\theta_{\mathrm{e}}}$ with a learnable parameter set $\theta_{\mathrm{e}}$ and a prediction network $g_{\theta_{\mathrm{clf}}}$ with a learnable parameter set $\theta_{\mathrm{clf}}$. The model $f$ is described as $f_{\theta}(x) = g_{\theta_{\mathrm{clf}}} (h_{\theta_{\mathrm{e}}}(x))$. The representation learning network acts as an encoder. The prediction network is a single linear layer followed by a softmax.

\subsection{Training} \label{subsec:train}
Each episode has $m$ batches of examples instantaneously drawn from data stream. For each task, our model is trained on $\mathcal{D}^{label}$ first, and then on $\mathcal{D}^{pool}$.  MAML consists of two optimization loops:

\subsubsection{Inner-Loop Optimization} 
Inner-loop algorithm performs task-specific tuning. We introduce data augmentations as the regularization term to improve intra-task generalization. The inner loop loss for training samples $\mathcal{D}_{i}^{label}$ at time step $i$ is
\begin{equation} \label{eqn:inner_label}
\small
    \mathcal{L}_{\mathbf{inner}}^{D_{i}^{label}}(\theta) = \sum_{ (x,y) \in D_{i}^{label}} [ w \mathcal{L}_{CE}(f_{\theta}(x),y ) + (1-w) \mathcal{L}_{CE}\big(f_{\theta}(\alpha(x)),y \big) ]
\end{equation} where $w$ denotes the relative weight and $\mathcal{L}_{CE}$ is the cross-entropy loss. 

\paragraph{Annotation Process} When the received training batches are unlabeled, we apply acquisition function $a(\mathcal{D}_{i}^{pool}, m \cdot n_{a})$ to select informative data points for annotation. The selection size per batch for annotation $n_{a} = \lceil \frac{b\cdot B_{A}}{|\mathcal{D}^{pool}|} \rceil$ where $b$ is the batch size and $B_A$ is the annotation budget. \\

\noindent Then, the inner-loop loss for newly annotated training batches $\mathcal{D}^{new}_i$ is
\begin{equation} \label{eqn:inner_unlabel}
\small
    \mathcal{L}_{\mathbf{inner}}^{\mathcal{D}^{new}_i}(\theta) = \sum_{ (x,y) \in \mathcal{D}^{new}_i} [w  \mathcal{L}_{CE}\big(f_{\theta}(x),y\big)  + (1-w) \mathcal{L}_{CE}\big(f_{\theta}(\mathcal{A}(x)),f_{\theta}(\alpha(x))\big)]
\end{equation} We use different inner loop loss for already-labeled data $\mathcal{D}_{i}^{label}$ and newly-labeled data $ \mathcal{D}^{new}_i$. The reason is that newly labeled data may contains more accurate and up-to-date label information. Then, our model performs SGD on parameter set $\theta_{\mathrm{clf}}$ with learning rate $\alpha$ as
\begin{equation}
    \tilde{\theta}_{\mathrm{clf}}=\theta_{\mathrm{clf}} - \alpha \nabla_{\theta} \mathcal{L}_{\mathbf{inner}}(\theta)
\end{equation}

\paragraph{Memory Sample Selection} 
We dynamically update memory buffer $\mathcal{M}$ using reservoir sampling to ensure generalization while avoiding overfitting. Reservoir sampling \cite{DBLP:conf/iclr/RiemerCALRTT19} randomly selects a fixed number of training samples without knowing the total number of samples in advance. We use reservoir sampling to select $n_s$ examples per class on the incoming data stream $\mathcal{D}^{label}_i \bigcup \mathcal{D}^{new}_i$ with an equal selection probability for all data seen so far. Note that the current label space $\mathcal{Y}_i$ might overlap with previous label spaces. We automatically update memory samples for $\mathcal{Y}_i$ to maintain a fixed amount of memory samples per class.

\begin{algorithm}[htbp]
\small
\caption{Continual Active Learning}
\label{algo:train-cal}
\begin{algorithmic}[1]
\Require Training set $\mathcal{D}^{train} = \mathcal{D}^{label} \cup \mathcal{D}^{pool}$, acquisition function $a(\cdot)$, inner-loop learning rate $\alpha$, outer-loop learning rate $\beta$, memory samples per class $n_s$, annotation budget per task $B_A$
\Ensure  Trained model with parameters  $ \theta = \theta_{\mathrm{e}} \cup \theta_{\mathrm{clf}}$ and memory buffer $\mathcal{M}$
\For{$i = 1, 2, \ldots$}
    \State Receiving $m$ batches of examples, $\mathcal{D}_{i}^{train}$, from the stream
    \If{$\mathcal{D}_{i}^{train}$ is from an unseen task}
        \State Compute selection size per batch for annotation as $n_{a} = \lceil \frac{b \cdot B_{A}}{|\mathcal{D}^{pool}|} \rceil$
    \EndIf
    \State \textbf{[Inner Loop]}
    \Comment{Trained on Labeled data first for each task}
    \If{$\mathcal{D}_{i}^{train} \subseteq \mathcal{D}^{label}$}
        \State $\mathcal{D}_{i}^{label} \gets \mathcal{D}_{i}^{train}$ and apply \textit{weak augmentation} on samples from $\mathcal{D}_{i}^{label}$
        \State Perform SGD on $\theta_{\mathrm{clf}}$ to minimize Eqn. \ref{eqn:inner_label}
    \ElsIf{$\mathcal{D}_{i}^{train} \subseteq \mathcal{D}^{pool}$}
        \State $\mathcal{D}_{i}^{pool} \leftarrow \mathcal{D}_{i}^{train}$ and do \textbf{annotation} \underline{per batch} as 
        $\mathcal{D}_{i}^{new} \xleftarrow{\text{label}} a(\mathcal{D}_{i}^{pool}, m \cdot n_{a})$
        \State Apply both \textit{weak and strong augmentations} on samples from $\mathcal{D}_{i}^{new}$
        \State Perform SGD on $\theta_{\mathrm{clf}}$ to minimize Eqn. \ref{eqn:inner_unlabel}
    \EndIf
    \State Select memory sample using \textit{ReservoirSampling}$(\mathcal{D}^{label}_i \bigcup \mathcal{D}^{new}_i, \mathcal{M})$ and update $\mathcal{M}$
    \State \textbf{[Outer Loop]}
    \State Read \textbf{ALL} examples from $\mathcal{M}$ and apply \textit{strong augmentation}
    \State Perform Adam update on $\theta$ to minimize Eqn. \ref{eqn:outer}
    \If{all training data are seen}
        \State \textbf{Stop training}
    \EndIf
\EndFor
\end{algorithmic}
\end{algorithm}

\subsubsection{Outer-Loop Optimization} 
Outer-loop algorithm optimizes initial parameter set $\theta$ to a setting that $f$ can effectively adapt to $\mathcal{D}_{1:t}$ via a few gradient updates. Due to memory constraints, we only have a few amount of samples from the past. The model reads all examples from memory buffer $\mathcal{M}$. Then, the outer-loop objective is to have $ \tilde{\theta} = \theta_{\mathrm{e}} \cup \tilde{\theta}_{\mathrm{clf}} $ generalize well on $\mathcal{D}_{1:t}$ using $\mathcal{M}$, shown in Eqn. \ref{eqn:cal_obj},
\begin{equation}
\begin{split}
    \mathcal{L}^{\mathcal{M}}_{\mathbf{meta}}(\tilde{\theta}) & = \sum_{(x,y) \in \mathcal{M}} [\mathcal{L}_{CE} (f_{\tilde{\theta}}(x),y)] \\
    & = \sum_{(x,y) \in \mathcal{M}} [\mathcal{L}_{CE}(g_{\tilde{\theta}_{\mathrm{clf}}} (h_{\theta_{\mathrm{e}}}(x)),y)]  \\
\end{split}
\end{equation} To improve both \textit{intra-} and \textit{inter-task generalization}, we employ strong augmentation to memory samples in meta-objective as
\begin{equation} \label{eqn:outer}
\mathcal{L}^{\mathcal{M}}_{\mathbf{meta}}(\tilde{\theta}) = \sum_{(x,y) \in \mathcal{M}} [\mathcal{L}_{CE}(g_{\tilde{\theta}_{\mathrm{clf}}} (h_{\theta_{\mathrm{e}}}(\mathcal{A}(x)),y)]  \\
\end{equation} To reduce the complexity of the second-order computation in outer loop, we use first-order approximation, namely FOMAML. The outer-loop optimization process is
\begin{equation}
    \theta \leftarrow \theta - \beta \nabla_{\tilde{\theta}} \mathcal{L}^{\mathcal{M}}_{\mathbf{meta}}(\tilde{\theta})  
\end{equation} where $\beta$ is the outer-loop learning rate. 
Algorithm \ref{algo:train-cal} outlines the complete training procedure.

\subsection{Testing}
The model randomly samples $m$ batches of examples drawn from $\mathcal{M}$ as support set $S$ and performs SGD on these samples to finetune parameter set $\theta_{\mathrm{clf}}$ with learning rate $\alpha$. The inner-loop loss at test is
\begin{equation}\label{eqn:inner_test}
\mathcal{L}_{\mathbf{inner}}^{S}(\theta) = \sum_{ (x,y) \in S}[\mathcal{L}_{CE}(f_{\theta}(x),y)] \\
\end{equation} where $S \subseteq \mathcal{M}$. And the optimization process is,
\begin{equation}
    \tilde{\theta}_{\mathrm{clf}} =\theta_{\mathrm{clf}} - \alpha \nabla_{\theta} \mathcal{L}_{\mathbf{inner}}^{S}(\theta)
\end{equation} Then, we output the predication using parameter set $\tilde{\theta} = \theta_{\mathrm{e}} \cup \tilde{\theta}_{\mathrm{clf}}$ on the a test sample $x_{\mathrm{test}}$ as
\begin{equation}
    \hat{y}_{\mathrm{test}} = f_{\tilde{\theta}}(x_{\mathrm{test}})= g_{\tilde{\theta}_{\mathrm{clf}}} (h_{\theta_{\mathrm{e}}}(x_{\mathrm{test}}))
\end{equation} 

\section{Experiments} \label{sec:exp}
\subsection{Datasets} We use the text classification benchmark
datasets from \citep{DBLP:conf/nips/ZhangZL15}, including AGNews (news classification; 4 classes), Yelp (sentiment analysis; 5 classes), Amazon (sentiment analysis; 5 classes), DBpedia (Wikipedia article classification; 14 classes) and Yahoo (questions and answers categorization; 10 classes). This collection contains 5 tasks from 4 different domains, which covers class- and domain-incremental learning in task sequence. We randomly sample 5 labeled instances per class, 10,000 unlabeled instances, and 7,600 test examples from each of the datasets. Following prior studies \citep{DBLP:conf/emnlp/WangMPC20, Holla2020MetaLearningWS, 10058177}, we concatenate training sets in 4 different orderings as shown in Table \ref{tab:orders}.

\begin{table}[h]
\caption{\label{tab:orders} Input Dataset Orders}
\begin{tabular}{@{}cl@{}}
\toprule
Index & Dataset Orders \\
\midrule
1 &Yelp $\rightarrow$ AGNews $\rightarrow$ DBpedia $\rightarrow$ Amazon $\rightarrow$ Yahoo \\
2 & DBpedia $\rightarrow$ Yahoo $\rightarrow$ AGNews $\rightarrow$ Amazon $\rightarrow$ Yelp \\
3 & Yelp $\rightarrow$ Yahoo $\rightarrow$ Amazon $\rightarrow$ DBpedia $\rightarrow$ AGNews \\
4 & AGNews $\rightarrow$ Yelp $\rightarrow$ Amazon $\rightarrow$ Yahoo $\rightarrow$ DBpedia \\
\botrule
\end{tabular}
\end{table}

\subsection{Implementation details} 
\label{sec:impl-details}
Our example encoder is a pretrained BERT$_{\mathrm{BASE}}$ model \citep{DBLP:conf/naacl/DevlinCLT19}. The parameter size of our model is 109M. For learning rates, $\alpha = 1e^{-3}$ and $\beta = 3e^{-5}$. The training batch size is 16. The number of mini-batches in each episode, $m$ = 5. The label budget $B_{A}$ = 2000 examples per task and the memory budget $B_\mathcal{M}$ is 5 per class, i.e., $n_s = 5$. For textual augmentation, we swap words randomly as the weak augmentation. And, we apply the combination of swapping words randomly, deleting words randomly and substituting words by WordNet’s synonym, as the strong augmentation.  We use \textit{nplug}\footnote{https://github.com/makcedward/nlpaug 1.1.10}, a Python package to implement augmentations. All models are executed on Linux platform with 8 Nvidia Tesla A100 GPU and 40 GB of RAM. All experiments are performed using PyTorch\footnote{https://pytorch.org/ 1.10.0+cu113} \citep{NEURIPS2019_9015}. 

\subsection{Baselines} 
\textbf{\textit{Baseline model:}}
\begin{enumerate}
    \item \textbf{MAML-SEQ}: Online FOMAML algorithm.
    \item \textbf{OML-ER} \citep{Holla2020MetaLearningWS}: OML with 5\% episodic experience replay rate\footnote{The default 1\% rate shows bad performance} + reservoir sampling.
    \item \textbf{C-MAML}\footnote{Being modified for NLP tasks} \citep{10.5555/3495724.3496696}: OML with Meta \& CL objective alignment + reservoir sampling.
    \item \textbf{Meta-CAL (ours)}: OML with Meta \& CL objective alignment + consistency regularization + reservoir sampling.
    \item \textbf{FULL}: Supervised C-MAML, trained on full datasets.
\end{enumerate}

\noindent \textbf{\textit{Memory sample selection:}}
\begin{enumerate}
    \item \textbf{Prototype} \citep{10058177}: Select representative samples that are closest to dynamically updated prototypes in representation space.
    \item \textbf{Ring Buffer} \citep{Chaudhry2019ContinualLW}: Use a 'First-In, First-Out' scheme to update buffer.
    \item \textbf{Reservoir Sampling} \citep{DBLP:conf/iclr/RiemerCALRTT19}: Randomly select data with an equal selection probability.
\end{enumerate}

\subsection{Evaluation Metrics} 
Based on the prior work \citep{10444954}, we use three comprehensive and widely-used metrics in CL, i.e., accuracy, backward transfer and forward transfer. Let $R_{k,i}$ be the macro-averaged accuracy evaluated on the test set of the $i$ task after sequentially learning $t$ tasks, \\

\noindent \textbf{Accuracy:}
\begin{equation*}
    \mathrm{ACC}_t = \frac{1}{t} \sum^{t}_{i=1} R_{t,i}
\end{equation*} Overall accuracy is the weighted average accuracy of all seen tasks $\mathcal{T}_{1:T}$. \\

\noindent \textbf{Backward transfer (\textit{stability} evaluation):}
\begin{equation*}
    \mathrm{BWT}_k = \frac{1}{k-1} \sum^{k-1}_{i=1} (R_{k,i} - R_{i,i})
\end{equation*} BWT measures how the updated parameters affect model performance on all previously seen tasks. \\

\noindent \textbf{Forward transfer (\textit{plasticity} evaluation):}
\begin{equation*}
    \mathrm{FWT}_k = \frac{1}{k-1} \sum^{k}_{i=2} (R_{i,i} - R_{0,i})
\end{equation*} FWT quantifies the average impact of all preceding tasks on the current $k$ task.

\subsection{Main Results}

In Table \ref{tab:full}, we compare various models with four AL strategies. These strategies include Random (denoted as RAND), Representative via KMeans (denoted as REP), Diversity via KMeans (denoted as DIV), and Uncertainty via Least-confidence (denoted as UNC). The label budget $B_{A}$ = 2000 examples per task. Each record is the average of three best results from five runs.

\begin{table}[h]
\caption{Average Accuracy of Four Training Set Orders}
\label{tab:full}
\begin{tabular*}{\textwidth}{@{\extracolsep{\fill}}lcccccc}
\toprule
\multirow{2}{*}{Method} & \multirow{2}{*}{Save Ratio\footnotemark[1]} & \multicolumn{4}{@{}c@{}}{AL} & \multirow{2}{*}{AVG.} \\\cmidrule{3-6}
& & RAND & REP & DIV & UNC &  \\
\midrule
MAML-SEQ & \multirow{2}{*}{100.0\%} & 8.35 \scalebox{0.8}{$\pm$ 5.1} & 7.63 \scalebox{0.8}{$\pm$ 4.8} & 7.23 \scalebox{0.8}{$\pm$ 4.1} & 6.15 \scalebox{0.8}{$\pm$ 3.9} & 7.34 \scalebox{0.8}{$\pm$ 0.9} \\
OML-ER & & 59.2 \scalebox{0.8}{$\pm$ 1.9} & 57.4 \scalebox{0.8}{$\pm$ 1.7} & 57.3 \scalebox{0.8}{$\pm$ 3.1} & 47.4 \scalebox{0.8}{$\pm$ 3.3} & 55.3 \scalebox{0.8}{$\pm$ 5.4} \\
C-MAML & & 63.0 \scalebox{0.8}{$\pm$ 2.9} & 60.9 \scalebox{0.8}{$\pm$ 2.8} & 59.9 \scalebox{0.8}{$\pm$ 1.0} & 50.4 \scalebox{0.8}{$\pm$ 2.1} & 58.6 \scalebox{0.8}{$\pm$ 5.6} \\
Meta-CAL (Ours) & & & & & \\
\quad w/. Prototype & 1.6\% & 57.6 \scalebox{0.8}{$\pm$ 5.0} & 57.8 \scalebox{0.8}{$\pm$ 4.4} & 57.8 \scalebox{0.8}{$\pm$ 4.7} & 42.0 \scalebox{0.8}{$\pm$ 6.2} & 53.8 \scalebox{0.8}{$\pm$ 7.9} \\
\quad w/. Ring buffer & (5 per class) & $\mathbf{66.9}$ \scalebox{0.8}{$\pm$ 2.4} & \textbf{66.3} \scalebox{0.8}{$\pm$ 2.2} & 64.8 \scalebox{0.8}{$\pm$ 2.5} & 50.7 \scalebox{0.8}{$\pm$ 3.6} & 62.2 \scalebox{0.8}{$\pm$ 7.7} \\
\quad w/. Reservoir\footnotemark[2] & & 66.5 \scalebox{0.8}{$\pm$ 1.9} & 65.9 \scalebox{0.8}{$\pm$ 1.9} & \textbf{64.9} \scalebox{0.8}{$\pm$ 1.3} & \textbf{53.5} \scalebox{0.8}{$\pm$ 2.7} & \textbf{62.7} \scalebox{0.8}{$\pm$ 6.2} \\
FULL\footnotemark[3] & & & & & & \underline{64.1 \scalebox{0.8}{$\pm$ 1.9}} \\
\bottomrule
\end{tabular*}
\footnotetext[1]{The ratio of saved samples to the total number of seen samples}
\footnotetext[2]{Default memory selection scheme}
\footnotetext[3]{Upper bound performance}
\end{table}

The result shows that our method yields comparable results to the FULL baseline, which is trained on more than 10,000 labeled samples per task. This result indicates our method can effectively select 2,000 informative samples from 10,000 unlabeled data to maximize model performance. Compared to other baselines with the same MAML framework, our method obtains the highest average accuracy among the four commonly-used AL strategies, demonstrating its robustness to different AL approaches. It also indicates Meta-CAL has a great ability of preventing catastrophic forgetting after sequentially learning five tasks. 

We also perform paired t-tests. Our model with default setting is significantly better than C-MAML with p-values $<$ 0.04 for all four AL strategies. We also compare AL strategies for the proposed model, Random is significantly better than other strategies with p-values $<$ 0.03. 

As for the memory sample selection schemes, Ring Buffer outperforms Reservoir Sampling by less than 1\% in RAND and REP. Whereas, Reservoir Sampling exhibits smaller standard deviations, indicating its robustness to training set orders. Prototype Sampling has the worst performance. We perform further analysis in \S \ref{subsubsec:memory_insight}.

\begin{table}[h]
\caption{Comparison of Different AL Strategies}
\label{tab:al}
\begin{tabular}{@{}llcc@{}}
\toprule
\multicolumn{2}{c}{AL} & Accuracy & AVG. \\
\midrule
\multirow{3}{*}{UNC} & Entropy-based & 50.9 \scalebox{0.8}{$\pm$ 2.7} & \multirow{3}{*}{53.5} \\
 & Least-confidence & 53.5 \scalebox{0.8}{$\pm$ 2.7} & \\
 & Margin-confidence & 56.2 \scalebox{0.8}{$\pm$ 4.7} & \\
\midrule
\multirow{2}{*}{DIV} & Mean-vector & 50.7 \scalebox{0.8}{$\pm$ 1.3} & \multirow{2}{*}{57.8} \\
 & KMeans & 64.9 \scalebox{0.8}{$\pm$ 1.3} & \\
\midrule
\multirow{2}{*}{REP} & Mean-vector & 57.9 \scalebox{0.8}{$\pm$ 2.1} & \multirow{2}{*}{61.9} \\
 & KMeans & 65.9 \scalebox{0.8}{$\pm$ 1.9} & \\
\midrule
RAND &  & \textbf{66.5} \scalebox{0.8}{$\pm$ 1.9} & \textbf{66.5} \\
\bottomrule
\end{tabular}
\footnotetext{We report the average accuracy of training sequence Yelp $\rightarrow$ AGNews $\rightarrow$ DBpedia $\rightarrow$ Amazon $\rightarrow$ Yahoo.}
\end{table}

Table \ref{tab:full} and Table \ref{tab:al} demonstrate that Random is the best AL strategy for our model. As shown in Table \ref{tab:al}, for both Diversity- and Representative-based AL strategies, KMeans is the most effective approach. Uncertainty-based methods comparatively perform poorly. CL models aim to revisit past samples to enhance generalization. However, replaying and annotating uncertain samples may not sufficiently improve generalization capabilities, which can hinder effective knowledge consolidation and consequently harms model performance.

\section{Further Analysis} \label{sec:fa}
We use the training set order\footnote{The performance on this order is close to the average performance.} Yelp $\rightarrow$ AGNews $\rightarrow$ DBpedia $\rightarrow$ Amazon $\rightarrow$ Yahoo to perform further analysis. In this section,  we denote RAND, REP, DIV and UNC as Random, Representative (KMeans), Diversity (KMeans) and Uncertainty (Least-confidence), respectively.

\subsection{Stability \& Plasticity} 
We test the performance of different AL strategies in terms of stability and plasticity.

\begin{figure*}[h]
\centering
\subfloat[RAND]{\includegraphics[width=1.5in]{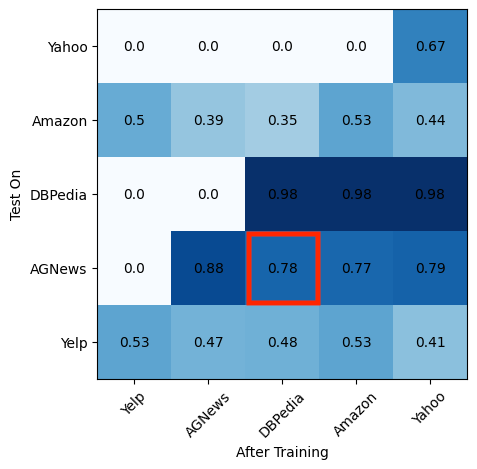}
\label{fig:rand}}
\subfloat[REP (KMeans)]{\includegraphics[width=1.5in]{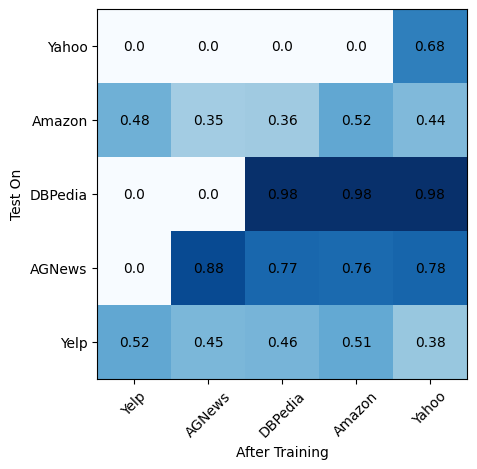}
\label{fig:rep}} \\
\subfloat[DIV (KMeans)]{\includegraphics[width=1.5in]{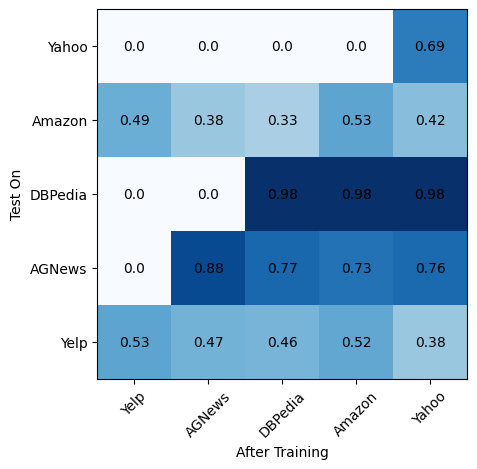}
\label{fig:div}}
\subfloat[UNC (Least-confidence)]{\includegraphics[width=1.5in]{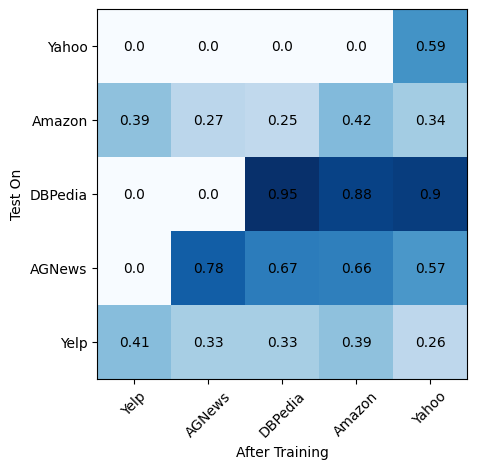}
\label{fig:unc}}
\caption{\label{fig:heatmap}Per task accuracy at different learning stages. The dark color indicates high accuracy. From left to right, the color for each task progressively fades away, indicating forgetting happens while learning more tasks. Note that Yelp and Amazon are from the same domain (sentiment analysis).UNC shows a lighter color compared to other AL methods, indicating a higher degree of forgetting. The red-outlined box shows the accuracy on AGNews (Task 2) after learning DBpedia (Task 3).}
\end{figure*}

\begin{figure}[h]
\centering
\includegraphics[width=0.9\textwidth]{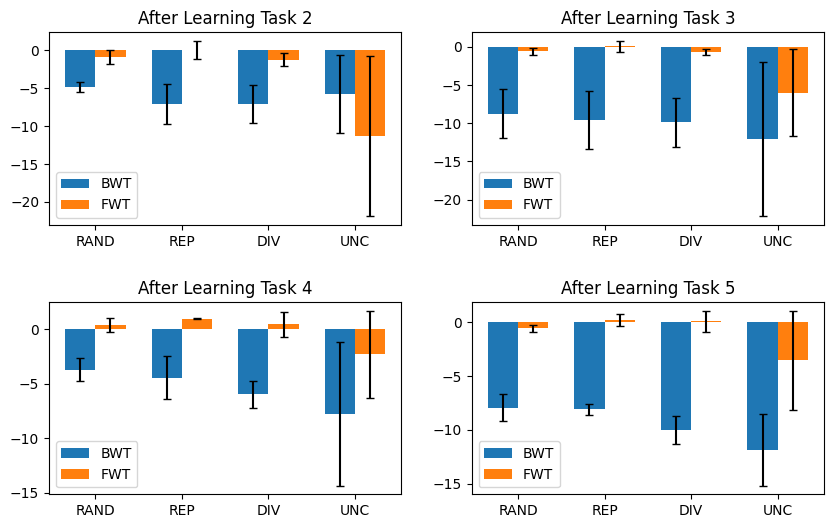}
\caption{\label{fig:eval} BWT and FWT for different AL strategies at each learning stage.}
\end{figure}

\noindent \textbf{Stability:} We employ the BWT metric as an indicator of catastrophic forgetting, evaluating the impact of the updated model parameters on the performance across all previously learned tasks. The negative BWT values indicate forgetting. As shown in Fig. \ref{fig:eval}, RAND shows the least forgetting, which indicates the best stability on past tasks. However, all methods exhibit a large forgetting after sequentially learning Task 3. Task 3 (DBpedia, 14 classes) has a relatively large label space compared to Task 1 (Yelp, 5 classes) and Task 2 (AGNews, 4 classes). As shown in Fig. \ref{fig:heatmap}, the accuracy on DBpedia exceeds the accuracy on other tasks. Since MAML learns high-quality reusable features for fast adaptation \citep{DBLP:conf/iclr/RaghuRBV20}, it tends to find shared representations that specifically benefit tasks with large label spaces. \\

\noindent \textbf{Plasticity:} A positive FWT score indicates the model's ability to leverage knowledge from previously learned tasks, facilitating zero-shot learning and efficient adaptation to the new task. As shown in Fig. \ref{fig:eval}, RAND, REP, and DIV show positive FWT values after training Task 4. This observation indicates that successful forward knowledge transfer from the previously learned tasks occurs. Fig. \ref{fig:heatmap} further supports that this transfer mainly occurs within the same domain, demonstrating effective \textit{domain-incremental learning} capabilities. Specifically, the proposed method leverages prior knowledge acquired from a familiar domain to facilitate efficient adaptation to new tasks within that domain. \\

\noindent Overall, RAND shows the best \textit{stability} while REP shows the best performance in \textit{plasticity}.

\subsection{Memory Insight} \label{subsubsec:memory_insight}
The level of generalization can be inferred from data dispersion. In this section, we investigate the effect of memory samples resulting from active learning and memory sample selection strategies. \\

\begin{figure*}[htbp]
\centering
\subfloat[After Learning Task 1]{\includegraphics[width=1.5in]{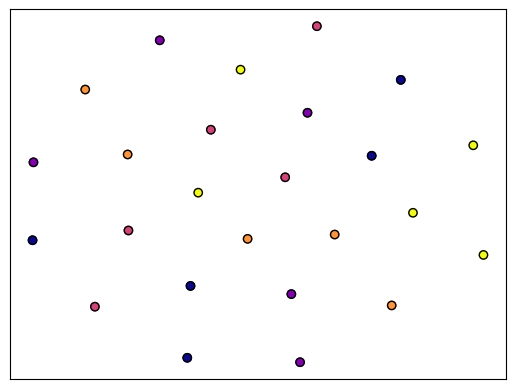}
\label{fig:1}}
\subfloat[After Learning Task 2]{\includegraphics[width=1.5in]{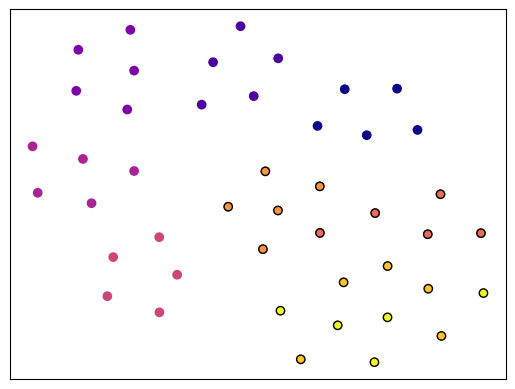}
\label{fig:2}}
\subfloat[After Learning Task 3]{\includegraphics[width=1.5in]{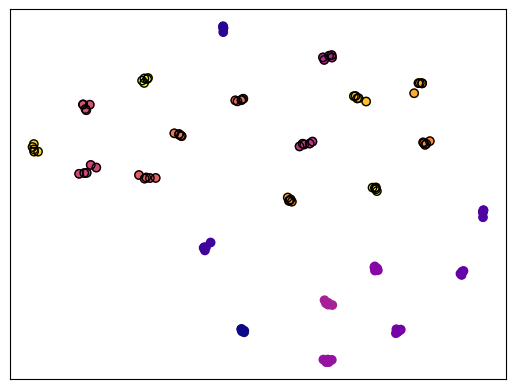}
\label{fig:3}} \\
\subfloat[After Learning Task 4]{\includegraphics[width=1.5in]{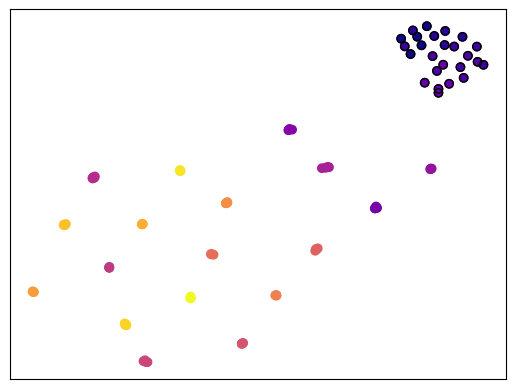}
\label{fig:4}}
\subfloat[After Learning Task 5]{\includegraphics[width=1.5in]{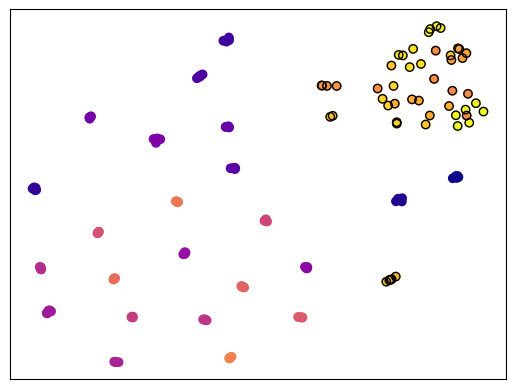}
\label{fig:5}}
\caption{\label{fig:insight-add} T-SNE visualization of memory samples at different learning stages using training set order Yelp $\rightarrow$ AGNews $\rightarrow$ DBpedia $\rightarrow$ Amazon $\rightarrow$ Yahoo. The black-circled data points belong to the last task. Data points with darker colors represent samples from earlier tasks, except (d) after learning Task 4. Task 4 has the same domain as Task 1. Hence, data points with darker colors are belong to the latest task in (d).}
\end{figure*}

\noindent \textbf{Different learning stages:} Fig. \ref{fig:insight-add} presents the T-SNE visualization of memory samples at different learning stages. As show in Figure \ref{fig:1} to \ref{fig:2}, when the proposed model learns a small number of tasks or encounters small label spaces, it focuses on ensuring intra-task generalization. However, as the number of seen tasks increases, the model shifts its focus towards ensuring inter-task generalization as shown in \ref{fig:4} to \ref{fig:5}. Consequently, we observe a phenomenon wherein the memory samples from the last tasks cluster together. \\

\begin{figure*}[htbp]
\centering
\subfloat[Reservoir-RAND(65.8\%)]{\includegraphics[width=1.4in]{fig/r-rand.png}
\label{fig:r-rand}} 
\subfloat[Reservoir-REP (64.7\%)]{\includegraphics[width=1.4in]{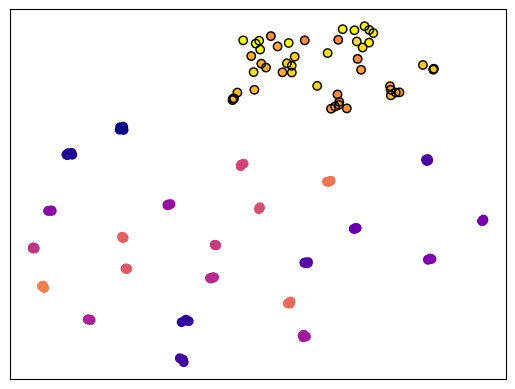}
\label{fig:r-rep}}
\subfloat[Reservoir-DIV (64.3\%)]{\includegraphics[width=1.4in]{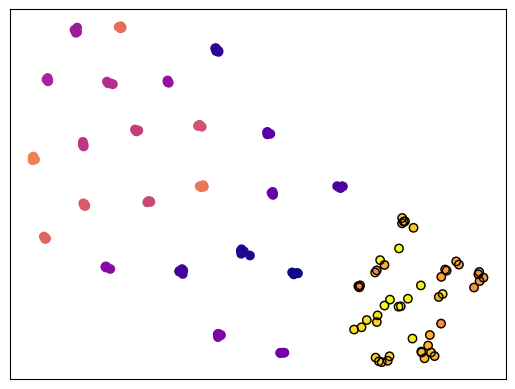}
\label{fig:r-div}} \\
\subfloat[Reservoir-UNC (53.8\%)]{\includegraphics[width=1.4in]{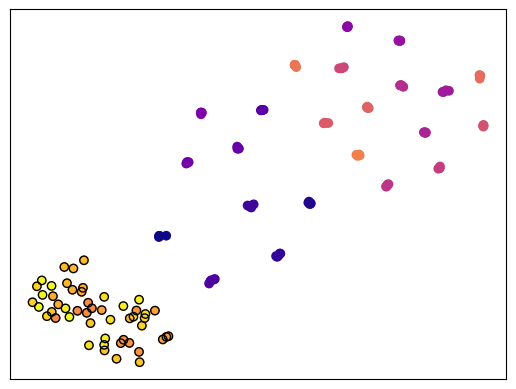}
\label{fig:r-unc}}
\subfloat[RingBuff.-RAND(65.4\%)]{\includegraphics[width=1.4in]{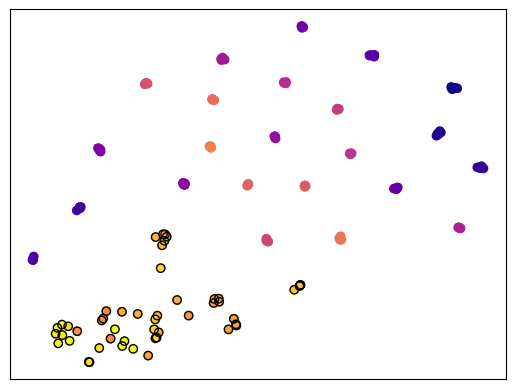}
\label{fig:rand-rf}}
\subfloat[Proto.-RAND(53.6\%)]{\includegraphics[width=1.4in]{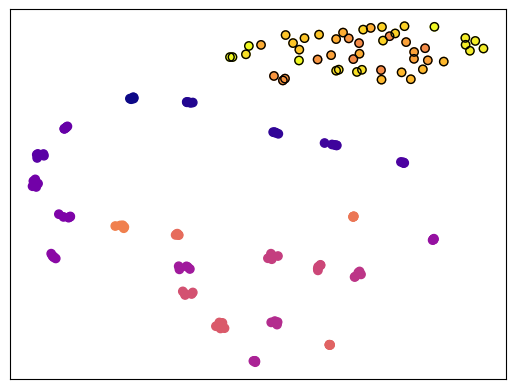}
\label{fig:rand-p}}
\caption{\label{fig:insight} T-SNE visualization of memory samples with different AL strategies. A equal dispersion of the data indicates a good memory representation. We provide accuracy for a better comparison. Data points with darker colors (purple, violet and pink) represent samples from earlier tasks. The black-circled data points belong to the last task. The clustering of memory samples from the last task suggests model focuses more on inter-task generalization rather than on intra-task generalization.}
\end{figure*}

\noindent \textbf{AL strategies:} As shown in Fig. \ref{fig:r-rand} to \ref{fig:r-unc}, memory data in RAND show a good data dispersion within the last task (i.e., \textit{intra-task generalization}) and across multiple tasks (i.e., \textit{inter-task generalization}), resulting in the best accuracy. In contrast, UNC in Fig. \ref{fig:r-unc} shows subpar performance in inter-task generalization. The result validates UNC's incapability to achieve generalization and preserve knowledge. Both REP and DIV exhibit lower accuracy compared to RAND, with DIV showing an obvious lower sparsity. Therefore, it is important to find an optimal balance between representativeness and diversity to ensure generalization. \\

\noindent \textbf{Memory sample selection methods:} In Fig. \ref{fig:r-rand}, \ref{fig:rand-rf} and \ref{fig:rand-p}, the choice of memory sample selection method also affects the model's performance. Prototype Sampling selects representative memory samples, potentially resulting in relatively less sparsity across multiple tasks. In contrast, Reservoir and Ring Buffer sampling strategies introduce randomness into the memory sample selection process, achieving better inter-task generalization. \\

\noindent Therefore, the randomness introduced by AL and memory sample selection methods can be beneficial in consolidating knowledge by ensuring \textbf{\textit{generalization}}. Furthermore, it is noteworthy that \textbf{\textit{inter-task confusion}} does not occur in the learned embedding space. This validates the superiority of our method in addressing confusion between old and new tasks.

\subsection{Effect of Augmentations} 

We conduct an ablation study to analyze textual augmentations and their key roles in Meta-CAL.  We examine the effect of inner-loop augmentation and outer loop augmentation in Table \ref{tab:abl-aug}. The meta samples in the outer-loop constrain the model behaviour. 

\begin{table}[htbp]
\renewcommand{\arraystretch}{1.3}
\caption{\label{tab:abl-aug} Ablation Study with Different Augmentation Modules
}
\centering
\begin{tabular}{l|cc}
\toprule
AL strategy:& 5-shot & 1-shot \\
RAND & $B_A = 2000$  & $B_A = 500$ \\
\midrule
Meta-CAL (Ours.) & 65.7 \scalebox{0.8}{$\pm$ 0.3} &  63.5 \scalebox{0.8}{$\pm$ 0.8} \\
\quad w/o. inner-loop aug.  & 65.8 \scalebox{0.8}{$\pm$ 0.4}  & 62.8 \scalebox{0.8}{$\pm$ 2.1}  \\
\quad w/o. outer-loop aug. &  58.7 \scalebox{0.8}{$\pm$ 2.3}  & 53.7 \scalebox{0.8}{$\pm$ 1.0}  \\
\quad w/o. aug. & 61.2 \scalebox{0.8}{$\pm$ 2.1} &  52.9 \scalebox{0.8}{$\pm$ 0.9} \\
\bottomrule
\end{tabular}
\end{table}

We hypothesize that if the meta-samples exhibit sufficient generalization capabilities, they can facilitate effective knowledge retention from prior tasks, consequently mitigating the issue of catastrophic forgetting. It is noteworthy that the meta-samples acquired through the combined effect of active learning acquisition and memory sample selection. We further enhance their generalization capabilities through data augmentation. The results validate this hypothesis, as we observe a significant gain by using outer-loop augmentation to enhance generalization. It improves accuracy by approximately 10\% when data availability and label budget are extremely limited. 

While inner-loop augmentation proves effective in extreme cases, it might not be as advantageous as outer-loop augmentation in non-extreme scenarios. Nevertheless, the combination of inner- and outer-loop augmentations still significantly improves accuracy.

\subsection{Annotation Budgets}
\begin{figure}[htbp]
\centering
\includegraphics[width=2.0in]{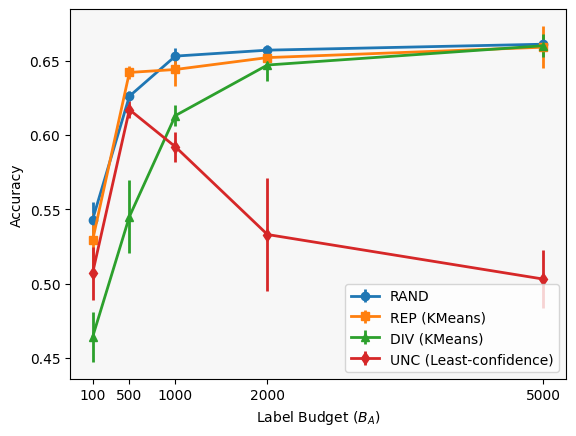}
\caption{Performance on different annotation budgets.}
\label{fig:labelbudget}
\end{figure}

As shown in Fig. \ref{fig:labelbudget}, in contrast to other AL strategies, increasing annotation budgets for UNC degrades model performance. This demonstrates that annotating uncertain samples is not beneficial for knowledge retention. REP achieves the best performance when the label budget is 500, while RAND outperforms other methods in most cases. Consequently, a certain degree of representativeness can aid in knowledge consolidation when annotation budget is extremely limited. In addition, our model achieves more than 62\% when the label budget is only 500 samples per task, suggesting its fast adaptation ability.

\subsection{Memory Budgets} 
We evaluate memory efficiency of our model. Table \ref{tab:memory_size} compares the performance of three models with the best average accuracy, i.e., our model with Reservoir sampling, our model with Ring Buffer and C-MAML. While all these models use the MAML framework, only the Meta-CAL models employ consistency regularization to enhance generalization. The result shows Meta-CAL models outperform C-MAML in all three cases. Notably, Meta-CAL w/. Reservoir can attain more than 50\% accuracy by saving only one sample per seen class. As a result, it also indicates that consistency regularization through data augmentation improves performance.

\begin{table}[htbp]
\renewcommand{\arraystretch}{1.3}
\caption{\label{tab:memory_size} Accuracy on different memory budgets. The AL strategy is RAND.}
\centering
\begin{tabular}{c|cc|c}
\toprule
Memory Budget & \multicolumn{2}{c|}{Meta-CAL}& \multirow{2}{*}{ C-MAML} \\
($n_s$ per class)  &Reservoir & Ring buffer & \\
\midrule
1  & 50.2 \scalebox{0.8}{$\pm$ 1.9}  &  46.5 \scalebox{0.8}{$\pm$ 3.4} & 47.7 \scalebox{0.8}{$\pm$ 6.0} \\		
5  & 65.7 \scalebox{0.8}{$\pm$ 0.3} & 66.4 \scalebox{0.8}{$\pm$ 0.5}  & 61.2 \scalebox{0.8}{$\pm$ 2.1} \\
10 &  67.6 \scalebox{0.8}{$\pm$ 0.1} & 68.0 \scalebox{0.8}{$\pm$ 0.8} & 63.9 \scalebox{0.8}{$\pm$ 0.6} \\
\bottomrule
\end{tabular}
\end{table}

\section{Conclusion}\label{sec13}
This paper considers a realistic continual learning scenario, namely few-shot continual active learning. To address this resource-constrained continual learning problems, we propose a novel method that employing meta-learning and active learning techniques, called \textit{Meta-Continual Active Learning}. Specifically, our model dynamically queries worthwhile unlabeled data for annotation and reformulate meta-objective with experience rehearsal and consistency regularization. In such a way, the proposed method is able to prevent catastrophic forgetting and improve generalization capability in a low-resource scenario. We conduct extensive experiments on benchmark text classification datasets in a 5-shot continual active learning setting. The results show the robustness of the proposed method. However, we only evaluate our method in a 5-shot case. In the future work, we can extend our evaluation to a more realistic scenario where the amount of labeled data is \{100, 1000, 5000\} and extend to other NLP tasks, e.g., language model training, text generation, and knowledge base enrichment. Furthermore, the annotation budget is allocated equally for each task. The annotation budget allocation strategies can be further studied.

\newpage
\bibliography{sn-bibliography}


\begin{thebibliography}{41}
\ifx \bisbn   \undefined \def \bisbn  #1{ISBN #1}\fi
\ifx \binits  \undefined \def \binits#1{#1}\fi
\ifx \bauthor  \undefined \def \bauthor#1{#1}\fi
\ifx \batitle  \undefined \def \batitle#1{#1}\fi
\ifx \bjtitle  \undefined \def \bjtitle#1{#1}\fi
\ifx \bvolume  \undefined \def \bvolume#1{\textbf{#1}}\fi
\ifx \byear  \undefined \def \byear#1{#1}\fi
\ifx \bissue  \undefined \def \bissue#1{#1}\fi
\ifx \bfpage  \undefined \def \bfpage#1{#1}\fi
\ifx \blpage  \undefined \def \blpage #1{#1}\fi
\ifx \burl  \undefined \def \burl#1{\textsf{#1}}\fi
\ifx \doiurl  \undefined \def \doiurl#1{\url{https://doi.org/#1}}\fi
\ifx \betal  \undefined \def \betal{\textit{et al.}}\fi
\ifx \binstitute  \undefined \def \binstitute#1{#1}\fi
\ifx \binstitutionaled  \undefined \def \binstitutionaled#1{#1}\fi
\ifx \bctitle  \undefined \def \bctitle#1{#1}\fi
\ifx \beditor  \undefined \def \beditor#1{#1}\fi
\ifx \bpublisher  \undefined \def \bpublisher#1{#1}\fi
\ifx \bbtitle  \undefined \def \bbtitle#1{#1}\fi
\ifx \bedition  \undefined \def \bedition#1{#1}\fi
\ifx \bseriesno  \undefined \def \bseriesno#1{#1}\fi
\ifx \blocation  \undefined \def \blocation#1{#1}\fi
\ifx \bsertitle  \undefined \def \bsertitle#1{#1}\fi
\ifx \bsnm \undefined \def \bsnm#1{#1}\fi
\ifx \bsuffix \undefined \def \bsuffix#1{#1}\fi
\ifx \bparticle \undefined \def \bparticle#1{#1}\fi
\ifx \barticle \undefined \def \barticle#1{#1}\fi
\bibcommenthead
\ifx \bconfdate \undefined \def \bconfdate #1{#1}\fi
\ifx \botherref \undefined \def \botherref #1{#1}\fi
\ifx \url \undefined \def \url#1{\textsf{#1}}\fi
\ifx \bchapter \undefined \def \bchapter#1{#1}\fi
\ifx \bbook \undefined \def \bbook#1{#1}\fi
\ifx \bcomment \undefined \def \bcomment#1{#1}\fi
\ifx \oauthor \undefined \def \oauthor#1{#1}\fi
\ifx \citeauthoryear \undefined \def \citeauthoryear#1{#1}\fi
\ifx \endbibitem  \undefined \def \endbibitem {}\fi
\ifx \bconflocation  \undefined \def \bconflocation#1{#1}\fi
\ifx \arxivurl  \undefined \def \arxivurl#1{\textsf{#1}}\fi
\csname PreBibitemsHook\endcsname

\bibitem[\protect\citeauthoryear{Andrychowicz et~al.}{2016}]{DBLP:conf/nips/AndrychowiczDCH16}
\begin{bchapter}
\bauthor{\bsnm{Andrychowicz}, \binits{M.}},
\bauthor{\bsnm{Denil}, \binits{M.}},
\bauthor{\bsnm{Colmenarejo}, \binits{S.G.}},
\bauthor{\bsnm{Hoffman}, \binits{M.W.}},
\bauthor{\bsnm{Pfau}, \binits{D.}},
\bauthor{\bsnm{Schaul}, \binits{T.}},
\bauthor{\bsnm{Freitas}, \binits{N.}}:
\bctitle{Learning to learn by gradient descent by gradient descent}.
In: \beditor{\bsnm{Lee}, \binits{D.D.}},
\beditor{\bsnm{Sugiyama}, \binits{M.}},
\beditor{\bsnm{Luxburg}, \binits{U.}},
\beditor{\bsnm{Guyon}, \binits{I.}},
\beditor{\bsnm{Garnett}, \binits{R.}} (eds.)
\bbtitle{Advances in Neural Information Processing Systems 29: Annual Conference on Neural Information Processing Systems 2016, December 5-10, 2016, Barcelona, Spain},
pp. \bfpage{3981}--\blpage{3989}
(\byear{2016})
\end{bchapter}
\endbibitem

\bibitem[\protect\citeauthoryear{Ayub and Fendley}{2022}]{DBLP:conf/nips/AyubF22}
\begin{bchapter}
\bauthor{\bsnm{Ayub}, \binits{A.}},
\bauthor{\bsnm{Fendley}, \binits{C.}}:
\bctitle{Few-shot continual active learning by a robot}.
In: \beditor{\bsnm{Koyejo}, \binits{S.}},
\beditor{\bsnm{Mohamed}, \binits{S.}},
\beditor{\bsnm{Agarwal}, \binits{A.}},
\beditor{\bsnm{Belgrave}, \binits{D.}},
\beditor{\bsnm{Cho}, \binits{K.}},
\beditor{\bsnm{Oh}, \binits{A.}} (eds.)
\bbtitle{Advances in Neural Information Processing Systems 35: Annual Conference on Neural Information Processing Systems 2022, NeurIPS 2022, New Orleans, LA, USA, November 28 - December 9, 2022}
(\byear{2022})
\end{bchapter}
\endbibitem

\bibitem[\protect\citeauthoryear{Adel et~al.}{2020}]{DBLP:conf/iclr/Adel0T20}
\begin{bchapter}
\bauthor{\bsnm{Adel}, \binits{T.}},
\bauthor{\bsnm{Zhao}, \binits{H.}},
\bauthor{\bsnm{Turner}, \binits{R.E.}}:
\bctitle{Continual learning with adaptive weights {(CLAW)}}.
In: \bbtitle{8th International Conference on Learning Representations, {ICLR} 2020, Addis Ababa, Ethiopia, April 26-30, 2020}
(\byear{2020})
\end{bchapter}
\endbibitem

\bibitem[\protect\citeauthoryear{Bachman et~al.}{2014}]{DBLP:conf/nips/BachmanAP14}
\begin{bchapter}
\bauthor{\bsnm{Bachman}, \binits{P.}},
\bauthor{\bsnm{Alsharif}, \binits{O.}},
\bauthor{\bsnm{Precup}, \binits{D.}}:
\bctitle{Learning with pseudo-ensembles}.
In: \beditor{\bsnm{Ghahramani}, \binits{Z.}},
\beditor{\bsnm{Welling}, \binits{M.}},
\beditor{\bsnm{Cortes}, \binits{C.}},
\beditor{\bsnm{Lawrence}, \binits{N.D.}},
\beditor{\bsnm{Weinberger}, \binits{K.Q.}} (eds.)
\bbtitle{Advances in Neural Information Processing Systems 27: Annual Conference on Neural Information Processing Systems 2014, December 8-13 2014, Montreal, Quebec, Canada},
pp. \bfpage{3365}--\blpage{3373}
(\byear{2014})
\end{bchapter}
\endbibitem

\bibitem[\protect\citeauthoryear{Beaulieu et~al.}{2020}]{DBLP:conf/ecai/BeaulieuFMLSCC20}
\begin{botherref}
\oauthor{\bsnm{Beaulieu}, \binits{S.}},
\oauthor{\bsnm{Frati}, \binits{L.}},
\oauthor{\bsnm{Miconi}, \binits{T.}},
\oauthor{\bsnm{Lehman}, \binits{J.}},
\oauthor{\bsnm{Stanley}, \binits{K.O.}},
\oauthor{\bsnm{Clune}, \binits{J.}},
\oauthor{\bsnm{Cheney}, \binits{N.}}:
In: {ECAI} 2020 - 24th European Conference on Artificial Intelligence, 29 August-8 September 2020, Santiago de Compostela, Spain, August 29 - September 8, 2020 - Including 10th Conference on Prestigious Applications of Artificial Intelligence {(PAIS} 2020).
Frontiers in Artificial Intelligence and Applications,
vol. 325,
pp. 992--1001.
{IOS} Press
(2020)
\end{botherref}
\endbibitem

\bibitem[\protect\citeauthoryear{Culotta and McCallum}{2005}]{DBLP:conf/aaai/CulottaM05}
\begin{bchapter}
\bauthor{\bsnm{Culotta}, \binits{A.}},
\bauthor{\bsnm{McCallum}, \binits{A.}}:
\bctitle{Reducing labeling effort for structured prediction tasks}.
In: \bbtitle{Proceedings, The Twentieth National Conference on Artificial Intelligence and the Seventeenth Innovative Applications of Artificial Intelligence Conference, July 9-13, 2005, Pittsburgh, Pennsylvania, {USA}},
pp. \bfpage{746}--\blpage{751}
(\byear{2005})
\end{bchapter}
\endbibitem

\bibitem[\protect\citeauthoryear{Chaudhry et~al.}{2019}]{Chaudhry2019ContinualLW}
\begin{botherref}
\oauthor{\bsnm{Chaudhry}, \binits{A.}},
\oauthor{\bsnm{Rohrbach}, \binits{M.}},
\oauthor{\bsnm{Elhoseiny}, \binits{M.}},
\oauthor{\bsnm{Ajanthan}, \binits{T.}},
\oauthor{\bsnm{Dokania}, \binits{P.K.}},
\oauthor{\bsnm{Torr}, \binits{P.H.S.}},
\oauthor{\bsnm{Ranzato}, \binits{M.}}:
Continual learning with tiny episodic memories.
ArXiv
\textbf{abs/1902.10486}
(2019)
\end{botherref}
\endbibitem

\bibitem[\protect\citeauthoryear{Chaudhry et~al.}{2019}]{DBLP:conf/iclr/ChaudhryRRE19}
\begin{bchapter}
\bauthor{\bsnm{Chaudhry}, \binits{A.}},
\bauthor{\bsnm{Ranzato}, \binits{M.}},
\bauthor{\bsnm{Rohrbach}, \binits{M.}},
\bauthor{\bsnm{Elhoseiny}, \binits{M.}}:
\bctitle{Efficient lifelong learning with {A-GEM}}.
In: \bbtitle{7th International Conference on Learning Representations, {ICLR} 2019, New Orleans, LA, USA, May 6-9, 2019}
(\byear{2019})
\end{bchapter}
\endbibitem

\bibitem[\protect\citeauthoryear{Chen et~al.}{2023}]{DBLP:conf/acl/ChenWS23}
\begin{bchapter}
\bauthor{\bsnm{Chen}, \binits{X.}},
\bauthor{\bsnm{Wu}, \binits{H.}},
\bauthor{\bsnm{Shi}, \binits{X.}}:
\bctitle{Consistent prototype learning for few-shot continual relation extraction}.
In: \bbtitle{Proceedings of the 61st Annual Meeting of the Association for Computational Linguistics (Volume 1: Long Papers), {ACL} 2023, Toronto, Canada, July 9-14, 2023},
pp. \bfpage{7409}--\blpage{7422}
(\byear{2023})
\end{bchapter}
\endbibitem

\bibitem[\protect\citeauthoryear{Das et~al.}{2023}]{das2023continual}
\begin{botherref}
\oauthor{\bsnm{Das}, \binits{A.M.}},
\oauthor{\bsnm{Bhatt}, \binits{G.}},
\oauthor{\bsnm{Bhalerao}, \binits{M.M.}},
\oauthor{\bsnm{Gao}, \binits{V.R.}},
\oauthor{\bsnm{Yang}, \binits{R.}},
\oauthor{\bsnm{Bilmes}, \binits{J.}}:
Continual Active Learning
(2023)
\end{botherref}
\endbibitem

\bibitem[\protect\citeauthoryear{Devlin et~al.}{2019}]{DBLP:conf/naacl/DevlinCLT19}
\begin{bchapter}
\bauthor{\bsnm{Devlin}, \binits{J.}},
\bauthor{\bsnm{Chang}, \binits{M.}},
\bauthor{\bsnm{Lee}, \binits{K.}},
\bauthor{\bsnm{Toutanova}, \binits{K.}}:
\bctitle{{BERT:} pre-training of deep bidirectional transformers for language understanding}.
In: \bbtitle{Proceedings of the 2019 Conference of the North American Chapter of the Association for Computational Linguistics: Human Language Technologies, {NAACL-HLT} 2019, Minneapolis, MN, USA, June 2-7, 2019, Volume 1 (Long and Short Papers)},
pp. \bfpage{4171}--\blpage{4186}
(\byear{2019})
\end{bchapter}
\endbibitem

\bibitem[\protect\citeauthoryear{de~Masson~d'Autume et~al.}{2019}]{DBLP:conf/nips/dAutumeRKY19}
\begin{bchapter}
\bauthor{\bsnm{Masson~d'Autume}, \binits{C.}},
\bauthor{\bsnm{Ruder}, \binits{S.}},
\bauthor{\bsnm{Kong}, \binits{L.}},
\bauthor{\bsnm{Yogatama}, \binits{D.}}:
\bctitle{Episodic memory in lifelong language learning}.
In: \beditor{\bsnm{Wallach}, \binits{H.M.}},
\beditor{\bsnm{Larochelle}, \binits{H.}},
\beditor{\bsnm{Beygelzimer}, \binits{A.}},
\beditor{\bsnm{d'Alch{\'{e}}{-}Buc}, \binits{F.}},
\beditor{\bsnm{Fox}, \binits{E.B.}},
\beditor{\bsnm{Garnett}, \binits{R.}} (eds.)
\bbtitle{Advances in Neural Information Processing Systems 32: Annual Conference on Neural Information Processing Systems 2019, NeurIPS 2019, December 8-14, 2019, Vancouver, BC, Canada},
pp. \bfpage{13122}--\blpage{13131}
(\byear{2019})
\end{bchapter}
\endbibitem

\bibitem[\protect\citeauthoryear{Finn et~al.}{2017}]{DBLP:conf/icml/FinnAL17}
\begin{bchapter}
\bauthor{\bsnm{Finn}, \binits{C.}},
\bauthor{\bsnm{Abbeel}, \binits{P.}},
\bauthor{\bsnm{Levine}, \binits{S.}}:
\bctitle{Model-agnostic meta-learning for fast adaptation of deep networks}.
In: \bbtitle{Proceedings of the 34th International Conference on Machine Learning, {ICML} 2017, Sydney, NSW, Australia, 6-11 August 2017}.
\bsertitle{Proceedings of Machine Learning Research},
vol. \bseriesno{70},
pp. \bfpage{1126}--\blpage{1135}
(\byear{2017})
\end{bchapter}
\endbibitem

\bibitem[\protect\citeauthoryear{Gupta et~al.}{2020}]{10.5555/3495724.3496696}
\begin{bchapter}
\bauthor{\bsnm{Gupta}, \binits{G.}},
\bauthor{\bsnm{Yadav}, \binits{K.}},
\bauthor{\bsnm{Paull}, \binits{L.}}:
\bctitle{La-maml: Look-ahead meta learning for continual learning}.
In: \bbtitle{Proceedings of the 34th International Conference on Neural Information Processing Systems}.
\bsertitle{NIPS'20}.
\bpublisher{Curran Associates Inc.},
\blocation{Red Hook, NY, USA}
(\byear{2020})
\end{bchapter}
\endbibitem

\bibitem[\protect\citeauthoryear{Huang et~al.}{2023}]{DBLP:conf/aaai/HuangC0CW23}
\begin{bchapter}
\bauthor{\bsnm{Huang}, \binits{B.}},
\bauthor{\bsnm{Chen}, \binits{Z.}},
\bauthor{\bsnm{Zhou}, \binits{P.}},
\bauthor{\bsnm{Chen}, \binits{J.}},
\bauthor{\bsnm{Wu}, \binits{Z.}}:
\bctitle{Resolving task confusion in dynamic expansion architectures for class incremental learning}.
In: \bbtitle{Thirty-Seventh {AAAI} Conference on Artificial Intelligence, {AAAI} 2023, Thirty-Fifth Conference on Innovative Applications of Artificial Intelligence, {IAAI} 2023, Thirteenth Symposium on Educational Advances in Artificial Intelligence, {EAAI} 2023, Washington, DC, USA, February 7-14, 2023},
pp. \bfpage{908}--\blpage{916}
(\byear{2023})
\end{bchapter}
\endbibitem

\bibitem[\protect\citeauthoryear{Han et~al.}{2020}]{DBLP:conf/acl/HanDGLLLSZ20}
\begin{bchapter}
\bauthor{\bsnm{Han}, \binits{X.}},
\bauthor{\bsnm{Dai}, \binits{Y.}},
\bauthor{\bsnm{Gao}, \binits{T.}},
\bauthor{\bsnm{Lin}, \binits{Y.}},
\bauthor{\bsnm{Liu}, \binits{Z.}},
\bauthor{\bsnm{Li}, \binits{P.}},
\bauthor{\bsnm{Sun}, \binits{M.}},
\bauthor{\bsnm{Zhou}, \binits{J.}}:
\bctitle{Continual relation learning via episodic memory activation and reconsolidation}.
In: \bbtitle{Proceedings of the 58th Annual Meeting of the Association for Computational Linguistics, {ACL} 2020, Online, July 5-10, 2020},
pp. \bfpage{6429}--\blpage{6440}
(\byear{2020})
\end{bchapter}
\endbibitem

\bibitem[\protect\citeauthoryear{Ho et~al.}{2023}]{10058177}
\begin{botherref}
\oauthor{\bsnm{Ho}, \binits{S.}},
\oauthor{\bsnm{Liu}, \binits{M.}},
\oauthor{\bsnm{Du}, \binits{L.}},
\oauthor{\bsnm{Gao}, \binits{L.}},
\oauthor{\bsnm{Xiang}, \binits{Y.}}:
Prototype-guided memory replay for continual learning.
IEEE Transactions on Neural Networks and Learning Systems,
1--11
(2023)
\end{botherref}
\endbibitem

\bibitem[\protect\citeauthoryear{Holla et~al.}{2020}]{Holla2020MetaLearningWS}
\begin{botherref}
\oauthor{\bsnm{Holla}, \binits{N.}},
\oauthor{\bsnm{Mishra}, \binits{P.}},
\oauthor{\bsnm{Yannakoudakis}, \binits{H.}},
\oauthor{\bsnm{Shutova}, \binits{E.}}:
Meta-learning with sparse experience replay for lifelong language learning.
ArXiv
\textbf{abs/2009.04891}
(2020)
\end{botherref}
\endbibitem

\bibitem[\protect\citeauthoryear{Javed and White}{2019}]{DBLP:conf/nips/JavedW19}
\begin{bchapter}
\bauthor{\bsnm{Javed}, \binits{K.}},
\bauthor{\bsnm{White}, \binits{M.}}:
\bctitle{Meta-learning representations for continual learning}.
In: \beditor{\bsnm{Wallach}, \binits{H.M.}},
\beditor{\bsnm{Larochelle}, \binits{H.}},
\beditor{\bsnm{Beygelzimer}, \binits{A.}},
\beditor{\bsnm{d'Alch{\'{e}}{-}Buc}, \binits{F.}},
\beditor{\bsnm{Fox}, \binits{E.B.}},
\beditor{\bsnm{Garnett}, \binits{R.}} (eds.)
\bbtitle{Advances in Neural Information Processing Systems 32: Annual Conference on Neural Information Processing Systems 2019, NeurIPS 2019, December 8-14, 2019, Vancouver, BC, Canada},
pp. \bfpage{1818}--\blpage{1828}
(\byear{2019})
\end{bchapter}
\endbibitem

\bibitem[\protect\citeauthoryear{Kirkpatrick et~al.}{2017}]{Kirkpatrick2017OvercomingCF}
\begin{barticle}
\bauthor{\bsnm{Kirkpatrick}, \binits{J.}},
\bauthor{\bsnm{Pascanu}, \binits{R.}},
\bauthor{\bsnm{Rabinowitz}, \binits{N.C.}},
\bauthor{\bsnm{Veness}, \binits{J.}},
\bauthor{\bsnm{Desjardins}, \binits{G.}},
\bauthor{\bsnm{Rusu}, \binits{A.A.}},
\bauthor{\bsnm{Milan}, \binits{K.}},
\bauthor{\bsnm{Quan}, \binits{J.}},
\bauthor{\bsnm{Ramalho}, \binits{T.}},
\bauthor{\bsnm{Grabska-Barwinska}, \binits{A.}},
\bauthor{\bsnm{Hassabis}, \binits{D.}},
\bauthor{\bsnm{Clopath}, \binits{C.}},
\bauthor{\bsnm{Kumaran}, \binits{D.}},
\bauthor{\bsnm{Hadsell}, \binits{R.}}:
\batitle{Overcoming catastrophic forgetting in neural networks}.
\bjtitle{Proceedings of the National Academy of Sciences}
\bvolume{114},
\bfpage{3521}--\blpage{3526}
(\byear{2017})
\end{barticle}
\endbibitem

\bibitem[\protect\citeauthoryear{Li and Hoiem}{2018}]{Li2018LearningWF}
\begin{barticle}
\bauthor{\bsnm{Li}, \binits{Z.}},
\bauthor{\bsnm{Hoiem}, \binits{D.}}:
\batitle{Learning without forgetting}.
\bjtitle{IEEE Transactions on Pattern Analysis and Machine Intelligence}
\bvolume{40},
\bfpage{2935}--\blpage{2947}
(\byear{2018})
\end{barticle}
\endbibitem

\bibitem[\protect\citeauthoryear{McCloskey and Cohen}{1989}]{McCloskey1989CatastrophicII}
\begin{barticle}
\bauthor{\bsnm{McCloskey}, \binits{M.}},
\bauthor{\bsnm{Cohen}, \binits{N.J.}}:
\batitle{Catastrophic interference in connectionist networks: The sequential learning problem}.
\bjtitle{Psychology of Learning and Motivation}
\bvolume{24},
\bfpage{109}--\blpage{165}
(\byear{1989})
\end{barticle}
\endbibitem

\bibitem[\protect\citeauthoryear{Mosqueira-Rey et~al.}{2022}]{mosqueira2022human}
\begin{botherref}
\oauthor{\bsnm{Mosqueira-Rey}, \binits{E.}},
\oauthor{\bsnm{Hern{\'a}ndez-Pereira}, \binits{E.}},
\oauthor{\bsnm{Alonso-R{\'\i}os}, \binits{D.}},
\oauthor{\bsnm{Bobes-Bascar{\'a}n}, \binits{J.}},
\oauthor{\bsnm{Fern{\'a}ndez-Leal}, \binits{{\'A}.}}:
Human-in-the-loop machine learning: a state of the art.
Artificial Intelligence Review,
1--50
(2022)
\end{botherref}
\endbibitem

\bibitem[\protect\citeauthoryear{Nichol et~al.}{2018}]{DBLP:journals/corr/abs-1803-02999}
\begin{botherref}
\oauthor{\bsnm{Nichol}, \binits{A.}},
\oauthor{\bsnm{Achiam}, \binits{J.}},
\oauthor{\bsnm{Schulman}, \binits{J.}}:
On first-order meta-learning algorithms
\textbf{abs/1803.02999}
(2018)
{\href{https://arxiv.org/abs/1803.02999}{{1803.02999}}}
\end{botherref}
\endbibitem

\bibitem[\protect\citeauthoryear{Netzer et~al.}{2011}]{37648}
\begin{bchapter}
\bauthor{\bsnm{Netzer}, \binits{Y.}},
\bauthor{\bsnm{Wang}, \binits{T.}},
\bauthor{\bsnm{Coates}, \binits{A.}},
\bauthor{\bsnm{Bissacco}, \binits{A.}},
\bauthor{\bsnm{Wu}, \binits{B.}},
\bauthor{\bsnm{Ng}, \binits{A.Y.}}:
\bctitle{Reading digits in natural images with unsupervised feature learning}.
In: \bbtitle{NIPS Workshop on Deep Learning and Unsupervised Feature Learning 2011}
(\byear{2011})
\end{bchapter}
\endbibitem

\bibitem[\protect\citeauthoryear{Paszke et~al.}{2019}]{NEURIPS2019_9015}
\begin{bchapter}
\bauthor{\bsnm{Paszke}, \binits{A.}},
\bauthor{\bsnm{Gross}, \binits{S.}},
\bauthor{\bsnm{Massa}, \binits{F.}},
\bauthor{\bsnm{Lerer}, \binits{A.}},
\bauthor{\bsnm{Bradbury}, \binits{J.}},
\bauthor{\bsnm{Chanan}, \binits{G.}},
\bauthor{\bsnm{Killeen}, \binits{T.}},
\bauthor{\bsnm{Lin}, \binits{Z.}},
\bauthor{\bsnm{Gimelshein}, \binits{N.}},
\bauthor{\bsnm{Antiga}, \binits{L.}},
\bauthor{\bsnm{Desmaison}, \binits{A.}},
\bauthor{\bsnm{Kopf}, \binits{A.}},
\bauthor{\bsnm{Yang}, \binits{E.}},
\bauthor{\bsnm{DeVito}, \binits{Z.}},
\bauthor{\bsnm{Raison}, \binits{M.}},
\bauthor{\bsnm{Tejani}, \binits{A.}},
\bauthor{\bsnm{Chilamkurthy}, \binits{S.}},
\bauthor{\bsnm{Steiner}, \binits{B.}},
\bauthor{\bsnm{Fang}, \binits{L.}},
\bauthor{\bsnm{Bai}, \binits{J.}},
\bauthor{\bsnm{Chintala}, \binits{S.}}:
\bctitle{Pytorch: An imperative style, high-performance deep learning library}.
In: \bbtitle{Advances in Neural Information Processing Systems 32},
pp. \bfpage{8024}--\blpage{8035}
(\byear{2019})
\end{bchapter}
\endbibitem

\bibitem[\protect\citeauthoryear{Perkonigg et~al.}{2021}]{Perkonigg2021ContinualAL}
\begin{botherref}
\oauthor{\bsnm{Perkonigg}, \binits{M.}},
\oauthor{\bsnm{Hofmanninger}, \binits{J.}},
\oauthor{\bsnm{Herold}, \binits{C.J.}},
\oauthor{\bsnm{Prosch}, \binits{H.}},
\oauthor{\bsnm{Langs}, \binits{G.}}:
Continual active learning using pseudo-domains for limited labelling resources and changing acquisition characteristics.
Machine Learning for Biomedical Imaging
(2021)
\end{botherref}
\endbibitem

\bibitem[\protect\citeauthoryear{Qin and Joty}{2022}]{DBLP:conf/acl/QinJ22}
\begin{bchapter}
\bauthor{\bsnm{Qin}, \binits{C.}},
\bauthor{\bsnm{Joty}, \binits{S.R.}}:
\bctitle{Continual few-shot relation learning via embedding space regularization and data augmentation}.
In: \bbtitle{Proceedings of the 60th Annual Meeting of the Association for Computational Linguistics (Volume 1: Long Papers), {ACL} 2022, Dublin, Ireland, May 22-27, 2022},
pp. \bfpage{2776}--\blpage{2789}
(\byear{2022})
\end{bchapter}
\endbibitem

\bibitem[\protect\citeauthoryear{Riemer et~al.}{2019}]{DBLP:conf/iclr/RiemerCALRTT19}
\begin{bchapter}
\bauthor{\bsnm{Riemer}, \binits{M.}},
\bauthor{\bsnm{Cases}, \binits{I.}},
\bauthor{\bsnm{Ajemian}, \binits{R.}},
\bauthor{\bsnm{Liu}, \binits{M.}},
\bauthor{\bsnm{Rish}, \binits{I.}},
\bauthor{\bsnm{Tu}, \binits{Y.}},
\bauthor{\bsnm{Tesauro}, \binits{G.}}:
\bctitle{Learning to learn without forgetting by maximizing transfer and minimizing interference}.
In: \bbtitle{7th International Conference on Learning Representations, {ICLR} 2019, New Orleans, LA, USA, May 6-9, 2019}
(\byear{2019})
\end{bchapter}
\endbibitem

\bibitem[\protect\citeauthoryear{Raghu et~al.}{2020}]{DBLP:conf/iclr/RaghuRBV20}
\begin{bchapter}
\bauthor{\bsnm{Raghu}, \binits{A.}},
\bauthor{\bsnm{Raghu}, \binits{M.}},
\bauthor{\bsnm{Bengio}, \binits{S.}},
\bauthor{\bsnm{Vinyals}, \binits{O.}}:
\bctitle{Rapid learning or feature reuse? towards understanding the effectiveness of {MAML}}.
In: \bbtitle{8th International Conference on Learning Representations, {ICLR} 2020, Addis Ababa, Ethiopia, April 26-30, 2020}
(\byear{2020})
\end{bchapter}
\endbibitem

\bibitem[\protect\citeauthoryear{Sohn et~al.}{2020}]{DBLP:conf/nips/SohnBCZZRCKL20}
\begin{bchapter}
\bauthor{\bsnm{Sohn}, \binits{K.}},
\bauthor{\bsnm{Berthelot}, \binits{D.}},
\bauthor{\bsnm{Carlini}, \binits{N.}},
\bauthor{\bsnm{Zhang}, \binits{Z.}},
\bauthor{\bsnm{Zhang}, \binits{H.}},
\bauthor{\bsnm{Raffel}, \binits{C.}},
\bauthor{\bsnm{Cubuk}, \binits{E.D.}},
\bauthor{\bsnm{Kurakin}, \binits{A.}},
\bauthor{\bsnm{Li}, \binits{C.}}:
\bctitle{Fixmatch: Simplifying semi-supervised learning with consistency and confidence}.
In: \bbtitle{Advances in Neural Information Processing Systems 33: Annual Conference on Neural Information Processing Systems 2020, NeurIPS 2020, December 6-12, 2020, Virtual}
(\byear{2020})
\end{bchapter}
\endbibitem

\bibitem[\protect\citeauthoryear{Shannon}{2001}]{10.1145/584091.584093}
\begin{barticle}
\bauthor{\bsnm{Shannon}, \binits{C.E.}}:
\batitle{A mathematical theory of communication}.
\bjtitle{SIGMOBILE Mob. Comput. Commun. Rev.}
\bvolume{5}(\bissue{1}),
\bfpage{3}--\blpage{55}
(\byear{2001})
\end{barticle}
\endbibitem

\bibitem[\protect\citeauthoryear{Schr{\"{o}}der et~al.}{2022}]{DBLP:conf/acl/SchroderNP22}
\begin{bchapter}
\bauthor{\bsnm{Schr{\"{o}}der}, \binits{C.}},
\bauthor{\bsnm{Niekler}, \binits{A.}},
\bauthor{\bsnm{Potthast}, \binits{M.}}:
\bctitle{Revisiting uncertainty-based query strategies for active learning with transformers}.
In: \bbtitle{Findings of the Association for Computational Linguistics: {ACL} 2022, Dublin, Ireland, May 22-27, 2022},
pp. \bfpage{2194}--\blpage{2203}
(\byear{2022})
\end{bchapter}
\endbibitem

\bibitem[\protect\citeauthoryear{van~de Ven et~al.}{2021}]{DBLP:conf/cvpr/Ven0T21}
\begin{bchapter}
\bauthor{\bsnm{Ven}, \binits{G.M.}},
\bauthor{\bsnm{Li}, \binits{Z.}},
\bauthor{\bsnm{Tolias}, \binits{A.S.}}:
\bctitle{Class-incremental learning with generative classifiers}.
In: \bbtitle{{IEEE} Conference on Computer Vision and Pattern Recognition Workshops, {CVPR} Workshops 2021, Virtual, June 19-25, 2021},
pp. \bfpage{3611}--\blpage{3620}
(\byear{2021})
\end{bchapter}
\endbibitem

\bibitem[\protect\citeauthoryear{Wu et~al.}{2024}]{wu2024meta}
\begin{bchapter}
\bauthor{\bsnm{Wu}, \binits{Y.}},
\bauthor{\bsnm{Huang}, \binits{L.-K.}},
\bauthor{\bsnm{Wang}, \binits{R.}},
\bauthor{\bsnm{Meng}, \binits{D.}},
\bauthor{\bsnm{Wei}, \binits{Y.}}:
\bctitle{Meta continual learning revisited: Implicitly enhancing online hessian approximation via variance reduction}.
In: \bbtitle{The Twelfth International Conference on Learning Representations}
(\byear{2024})
\end{bchapter}
\endbibitem

\bibitem[\protect\citeauthoryear{Wang et~al.}{2020}]{DBLP:conf/emnlp/WangMPC20}
\begin{bchapter}
\bauthor{\bsnm{Wang}, \binits{Z.}},
\bauthor{\bsnm{Mehta}, \binits{S.V.}},
\bauthor{\bsnm{P{\'{o}}czos}, \binits{B.}},
\bauthor{\bsnm{Carbonell}, \binits{J.G.}}:
\bctitle{Efficient meta lifelong-learning with limited memory}.
In: \bbtitle{Proceedings of the 2020 Conference on Empirical Methods in Natural Language Processing, {EMNLP} 2020, Online, November 16-20, 2020},
pp. \bfpage{535}--\blpage{548}
(\byear{2020})
\end{bchapter}
\endbibitem

\bibitem[\protect\citeauthoryear{Wang et~al.}{2022}]{wang-etal-2022-learning-robust}
\begin{bchapter}
\bauthor{\bsnm{Wang}, \binits{P.}},
\bauthor{\bsnm{Song}, \binits{Y.}},
\bauthor{\bsnm{Liu}, \binits{T.}},
\bauthor{\bsnm{Lin}, \binits{B.}},
\bauthor{\bsnm{Cao}, \binits{Y.}},
\bauthor{\bsnm{Li}, \binits{S.}},
\bauthor{\bsnm{Sui}, \binits{Z.}}:
\bctitle{Learning robust representations for continual relation extraction via adversarial class augmentation}.
In: \bbtitle{Proceedings of the 2022 Conference on Empirical Methods in Natural Language Processing},
pp. \bfpage{6264}--\blpage{6278}
(\byear{2022})
\end{bchapter}
\endbibitem

\bibitem[\protect\citeauthoryear{Wang et~al.}{2019}]{DBLP:conf/naacl/WangXYGCW19}
\begin{bchapter}
\bauthor{\bsnm{Wang}, \binits{H.}},
\bauthor{\bsnm{Xiong}, \binits{W.}},
\bauthor{\bsnm{Yu}, \binits{M.}},
\bauthor{\bsnm{Guo}, \binits{X.}},
\bauthor{\bsnm{Chang}, \binits{S.}},
\bauthor{\bsnm{Wang}, \binits{W.Y.}}:
\bctitle{Sentence embedding alignment for lifelong relation extraction}.
In: \bbtitle{Proceedings of the 2019 Conference of the North American Chapter of the Association for Computational Linguistics: Human Language Technologies, {NAACL-HLT} 2019, Minneapolis, MN, USA, June 2-7, 2019, Volume 1 (Long and Short Papers)},
pp. \bfpage{796}--\blpage{806}
(\byear{2019})
\end{bchapter}
\endbibitem

\bibitem[\protect\citeauthoryear{Wang et~al.}{2024}]{10444954}
\begin{botherref}
\oauthor{\bsnm{Wang}, \binits{L.}},
\oauthor{\bsnm{Zhang}, \binits{X.}},
\oauthor{\bsnm{Su}, \binits{H.}},
\oauthor{\bsnm{Zhu}, \binits{J.}}:
A comprehensive survey of continual learning: Theory, method and application.
IEEE Transactions on Pattern Analysis and Machine Intelligence,
1--20
(2024)
\end{botherref}
\endbibitem

\bibitem[\protect\citeauthoryear{Yoon et~al.}{2018}]{DBLP:conf/iclr/YoonYLH18}
\begin{bchapter}
\bauthor{\bsnm{Yoon}, \binits{J.}},
\bauthor{\bsnm{Yang}, \binits{E.}},
\bauthor{\bsnm{Lee}, \binits{J.}},
\bauthor{\bsnm{Hwang}, \binits{S.J.}}:
\bctitle{Lifelong learning with dynamically expandable networks}.
In: \bbtitle{6th International Conference on Learning Representations, {ICLR} 2018, Vancouver, BC, Canada, April 30 - May 3, 2018, Conference Track Proceedings}
(\byear{2018})
\end{bchapter}
\endbibitem

\bibitem[\protect\citeauthoryear{Zhang et~al.}{2015}]{DBLP:conf/nips/ZhangZL15}
\begin{bchapter}
\bauthor{\bsnm{Zhang}, \binits{X.}},
\bauthor{\bsnm{Zhao}, \binits{J.J.}},
\bauthor{\bsnm{LeCun}, \binits{Y.}}:
\bctitle{Character-level convolutional networks for text classification}.
In: \beditor{\bsnm{Cortes}, \binits{C.}},
\beditor{\bsnm{Lawrence}, \binits{N.D.}},
\beditor{\bsnm{Lee}, \binits{D.D.}},
\beditor{\bsnm{Sugiyama}, \binits{M.}},
\beditor{\bsnm{Garnett}, \binits{R.}} (eds.)
\bbtitle{Advances in Neural Information Processing Systems 28: Annual Conference on Neural Information Processing Systems 2015, December 7-12, 2015, Montreal, Quebec, Canada},
pp. \bfpage{649}--\blpage{657}
(\byear{2015})
\end{bchapter}
\endbibitem

\end{thebibliography}

\newpage
\appendix

\section{Supplemental Material}

\begin{table*}[htbp]
\small
\renewcommand{\arraystretch}{1.2}
\caption{\label{tab:full-f}Accuracy of Four Training Set Orders and Standard Deviation}
\begin{tabular}{l|l|ccccc}
\toprule
AL & CL Method & Order 1 & Order 2 & Order 3 & Order 4 & Average \\
\midrule
\multirow{6}{*}{RAND}& MAML-SEQ &$ 4.1 \pm 0.3 $&	$	10.1 \pm 0.2$&	$	14.8 \pm 3.6$&	$	4.4 \pm 3.9$&	$	8.4 \pm 5.1$ \\
& OML-ER & $60.2 \pm 3.0 $ &	$	59.7 \pm 2.1 $ &	$	56.4 \pm 2.7$ &	$	60.4 \pm 4.0	$ &	$59.2 \pm 1.9$ \\
& C-MAML &$61.2 \pm 2.1 $ &	$	65.1 \pm 2.8	$ &	$65.8 \pm 1.0 $ &	$	59.8 \pm 2.3$ &	$	63.0 \pm 2.9$ \\

& Ours.(PROT) &$54.7 \pm 1.8 $ &	$	61.8 \pm 1.6	$ &	$62.1 \pm 1.8 $ &	$	52.4 \pm 2.7$ &	$	57.6 \pm 4.9$ \\
& Ours.(RB) &$66.4 \pm 0.5  $ &	$	68.1 \pm 0.6  $ &	$	69.2 \pm 0.2 $ &	$	63.8 \pm 0.2	 $ &	$66.9 \pm 2.4$ \\
& Ours.(RESV) &$65.7 \pm 0.3	$ &	$68.1 \pm 1.2$ &	$	68.1 \pm 1.2$ &	$	64.2 \pm 0.6$ &	$66.5 \pm 1.9$ \\
\midrule

& MAML-SEQ & $10.3 \pm 7.1$&	$	12.2 \pm 1.0$&	$	6.6 \pm 2.7$&	$	1.4 \pm 0.1$&	$	7.63 \pm 4.8$ \\
& OML-ER & $55.5 \pm 3.1$&$56.4 \pm 1.2	$&	$58.2 \pm 3.4$&	$	59.3 \pm 3.2$&	$57.4 \pm 1.7 $ \\
REP& C-MAML & $58.3 \pm 0.7$&$62.0 \pm 2.6	$&	$64.4 \pm 1.8$&	$	58.9 \pm 3.1$&	$60.9 \pm 2.8 $ \\
(KMeans)& Ours.(PROT) & $58.0 \pm 2.8	$&	$62.3 \pm 1.3	$&	$59.2 \pm 3.3 $&	$	51.8 \pm 0.6 $&	$	57.8 \pm 4.4$ \\
& Ours.(RB) & $65.3 \pm 1.0 $&	$	67.3 \pm 0.6 $&	$	68.7 \pm 1.4 $&	$	63.7 \pm 1.5	$&	$66.3 \pm 2.2$ \\
& Ours.(RESV) & $65.2 \pm 0.5 $&$67.5 \pm 0.9   $&	$67.3 \pm 0.2$&	$	63.5 \pm 1.2$&	$65.9 \pm 1.9$ \\

\midrule 
& MAML-SEQ & $ 7.3 \pm 2.2 $&	$	9.6 \pm 1.6$&	$	10.5 \pm 6.3$&	$	1.5 \pm 0.1$&	$	7.23 \pm 4.1$ \\
& OML-ER & $54.8 \pm 1.8 $&	$	54.5 \pm 1.2 $&	$	59.1 \pm 2.8 $&	$	60.8 \pm 3.1 $&	$	57.3 \pm 3.1$ \\
DIV & C-MAML &$60.9 \pm 1.7$&$	59.1 \pm 2.1	$&$60.6 \pm 1.7	$&$59.1 \pm 2.1$&$	59.9 \pm 1.0$\\
(KMeans) & Ours.(PROT) & $55.5 \pm 3.2 $&$	63.4 \pm 1.4	$&$59.7 \pm 1.1$&$	52.6 \pm 4.6	$&$57.8 \pm 4.7$ \\
& Ours.(RB) & $63.7 \pm 0.4 $&$	66.8 \pm 1.5	$&$66.8 \pm 0.3 $&$	61.7 \pm 2.1$&$	64.8 \pm 2.5$ \\
& Ours.(RESV) & $64.7 \pm 1.1$&	$	65.7 \pm 0.6$&	$	66.1 \pm 0.2$&	$	63.1 \pm 2.3$&	$64.9 \pm 1.3$ \\

\midrule

& MAML-SEQ & $4.2 \pm 2.0$ &	$9.2 \pm 1.8$ &$9.6 \pm 4.3$ &	$1.6 \pm 0.4$&	$6.15 \pm 3.9$ \\
& OML-ER & $47.1 \pm 1.5$&	$51.9 \pm 2.8$&$	46.7 \pm 7.6$&$	44.0 \pm 1.5$ & $	47.4 \pm 3.3$ \\
UNC& C-MAML & $ 50.9 \pm 3.3	$&	$49.2 \pm 1.5	$&	$53.1 \pm 2.5$&	$	48.4 \pm 0.4$&	$	50.4 \pm 2.1 $\\
(L-CONF)& Ours.(PROT) & $37.5 \pm 6.5 $&	$	49.5 \pm 0.9$&	$	44.6 \pm 6.0	$&	$36.3 \pm 3.5	$&	$42.0 \pm 6.2$ \\
& Ours.(RB) & $47.4 \pm 3.6 $&	$	53.9 \pm 1.1 $&	$ 53.7 \pm 2.0 $&	$	47.8 \pm 1.4	$&	$50.7 \pm 3.6$ \\
& Ours.(RESV) &$53.3 \pm 3.8 $&	$	54.4 \pm 0.5	$&	$56.3 \pm 1.3	$&	$49.8 \pm 0.8	$&$53.5 \pm 2.7$ \\
\midrule
REP(Means) & \multirow{4}{*}{Ours.(RESV)} & $57.5 \pm 0.9$ &  $ 58.1 \pm 0.4 $ & $ 60.6 \pm 1.2 $ & $ 55.5 \pm 1.0 $ & $ 57.9 \pm 2.1 $ \\
DIV(Means) &  & $50.0 \pm 1.5$ & $ 50.5 \pm 2.8 $ & $ 52.5 \pm 0.4 $ & $ 49.6 \pm 4.7 $ & $ 50.7 \pm 1.3 $  \\

UNC(MARG) & & $58.4 \pm 1.8$ & $ 58.0 \pm 0.9 $ & $ 59.1 \pm 2.1 $ & $ 49.2 \pm 1.4 $ & $ 56.2 \pm 4.7 $  \\

UNC(ENT) & & $51.2 \pm 0.3$ &  $ 49.2 \pm 1.4 $ & $ 54.6 \pm 2.3 $ & $ 48.5 \pm 4.1 $ & $ 50.9 \pm 2.7 $ \\
\midrule

FULL & C-MAML & $ 64.5 \pm 1.4 $  & $ 62.2 \pm 1.3 $ &$ 66.5 \pm 1.0 $& $ 63.2 \pm 1.7 $& $ 64.1 \pm 1.7 $  \\
\bottomrule
\end{tabular}
\end{table*}





\end{document}